\documentclass[pdflatex,sn-basic]{sn-jnl}

\usepackage{natbib}
\usepackage{booktabs}       
\usepackage{amsfonts}       
\usepackage{nicefrac}       
\usepackage{microtype}      
\usepackage{xcolor}         
\usepackage{amsmath,amssymb,bm}
\usepackage{subcaption}
\usepackage{float}
\usepackage{graphicx}
\usepackage{tablefootnote}
\usepackage{adjustbox}
\usepackage{mathtools}
\usepackage{multirow}
\usepackage[capitalize]{cleveref}

\jyear{2021}%

\theoremstyle{thmstyleone}%
\theoremstyle{thmstyletwo}%

\theoremstyle{thmstylethree}%

\begin{document}

\title[Heterogeneous Multi-Task Gaussian Cox Processes]{Heterogeneous Multi-Task Gaussian Cox Processes}


\author[1,2]{\fnm{Feng} \sur{Zhou}}\email{feng.zhou@ruc.edu.cn}

\author[3]{\fnm{Quyu} \sur{Kong}}\email{quyu.kong@uts.edu.au}

\author[2,4]{\fnm{Zhijie} \sur{Deng}}\email{zhijied@sjtu.edu.cn}

\author[5]{\fnm{Fengxiang} \sur{He}}\email{fengxiang.f.he@gmail.com}

\author[2]{\fnm{Peng} \sur{Cui}}\email{cuip22@mails.tsinghua.edu.cn}

\author*[2]{\fnm{Jun} \sur{Zhu}}\email{dcszj@tsinghua.edu.cn}

\affil[1]{\orgdiv{Center for Applied Statistics and School of Statistics}, \orgname{Renmin University of China}}

\affil[2]{\orgdiv{Dept. of Comp. Sci. \& Tech., BNRist Center, THU-Bosch Joint ML Center}, \orgname{Tsinghua University}}

\affil[3]{\orgdiv{Data Science Institute}, \orgname{University of Technology Sydney}}

\affil[4]{\orgdiv{Qing Yuan Research Institute}, \orgname{Shanghai Jiao Tong University}}

\affil[5]{\orgdiv{JD Explore Academy}, \orgname{JD.com Inc}}


\abstract{This paper presents a novel extension of multi-task Gaussian Cox processes for modeling multiple heterogeneous correlated tasks jointly, e.g., classification and regression, via multi-output Gaussian processes (MOGP). A MOGP prior over the parameters of the dedicated likelihoods for classification, regression and point process tasks can facilitate sharing of information between heterogeneous tasks, while allowing for nonparametric parameter estimation. To circumvent the non-conjugate Bayesian inference in the MOGP modulated heterogeneous multi-task framework, we employ the data augmentation technique and derive a mean-field approximation to realize closed-form iterative updates for estimating model parameters. We demonstrate the performance and inference on both 1D synthetic data as well as 2D urban data of Vancouver. }

\keywords{heterogeneous correlation, multi-task learning, Cox process, multi-output Gaussian processes, conditionally conjugate}



\maketitle

\section{Introduction}\label{intro}

Inhomogeneous Poisson process data defined on a continuous spatio-temporal domain has attracted immense attention recently in a wide variety of applications, including reliability analysis in manufacturing systems~\citep{soleimani2017scalable}, event capture in sensing regions~\citep{mutny2021no}, crime prediction in urban area~\citep{shirota2017space} and disease diagnosis based on medical records~\citep{lasko2014efficient}. The reliable training of an inhomogeneous Poisson process model critically relies on a large amount of data to avoid overfitting, especially when modeling high-dimensional point processes. However, one challenge is that the available training data is routinely sparse or even partially missing in specific applications. Taking manufacturing failure and healthcare analysis as motivating examples: the modern manufacturing machines are reliable and sparsely fail; the individuals with healthy constitution will not visit hospital very often. The data missing problems also arise, e.g., the event location capture is intermittent for sensing systems because of weather or other related barriers. To handle data sparse/missing problems, the correlation between multiple tasks can be exploited to facilitate sharing of information between all tasks to improve the generalization capabilities, forming a multi-task learning paradigm. 



A popular approach to modeling multi-task inhomogeneous Poisson processes is to use Gaussian process (GP)~\citep{williams2006gaussian} based Bayesian framework to induce correlation among tasks. This kind of multi-task inhomogeneous Poisson processes are also called multi-task Cox processes~\citep{moller1998log}. Multi-task Cox processes have been investigated extensively in recent years, e.g., hierarchical-GP based version~\citep{lian2015multitask} and multi-output Gaussian processes (MOGP) based versions~\citep{aglietti2019efficient,jahani2021multioutput}. Yet to our knowledge, all the aforementioned works focus on \emph{homogeneous} multi-task Cox processes learning, i.e., all correlated tasks are exclusively point process tasks. It is not free to apply them to the more general \emph{heterogeneous} multi-task scenarios where correlated tasks include other types of tasks except Cox processes. Take the urban data of Vancouver in \cref{fig3} as a motivating example where we have three types of tasks: employment income (regression), education degree (classification), theft of vehicle (Cox process) and non-market house (Cox process). When the crime data is missing in certain areas of the city, training on this single task is prone to overfitting since the model may try to fit the available data too closely, leading to inaccurate predictions or poor generalization to unseen data. Can we leverage the information of employment income, education degree and non-market housing to assist the prediction of crime rate in the missing areas? Or, can we make use of income, education and crime to help predict the number of non-market housing projects in certain missing areas? 
Based on our knowledge, only a few heterogeneous frameworks exist, such as \citet{moreno2018heterogeneous}. However, \citet{moreno2018heterogeneous} discretized the point process task into Poisson distribution problems and does not preserve conjugate operations. 
To make further progress, we generalize the homogeneous multi-task Cox processes to the heterogeneous setup using MOGP to enable the transfer of knowledge between supervised (regression and classification) and unsupervised tasks (Cox processes).

Most existing Cox process works focus on the log Gaussian Cox process (LGCP)~\citep{moller1998log} where a GP function is passed through an exponential link function to model the positive intensity rate. Due to the nonconjugacy between point process likelihood and GP prior, practitioners 
need to apply Markov chain Monte Carlo (MCMC)~\citep{neal1993probabilistic} or variational inference~\citep{blei2017variational} methods to infer the posterior distribution of model parameters. 
For MCMC, the specialised MCMC algorithms, such as Metropolis-adjusted Langevin algorithm (MALA)~\citep{moller1998log,besag1994discussion}, as well as the probabilistic programming languages based on MCMC~\citep{wood2014new} where one does not need to write a sampler by hand, can be used for sampling from the posterior of intensity function. 
Although MCMC provides the guarantee of asymptotic consistency, this accuracy comes at the expense of a high computational cost. 
On the contrary, variational inference can be faster than MCMC, although it induces approximation error. 
For the efficiency reason, we focus on variational inference in this work. 
For variational inference, 
a Gaussian variational posterior is typically assumed to render the evidence lower bound (ELBO) tractable~\citep{dezfouli2015scalable,lloyd2015variational}. While this variational inference method is quite generic, it can exhibit low efficiency (although it is still faster than MCMC)~\citep{wenzel2019efficient}, exposing opportunities for improvement. It is worth noting that the same problem also occurs in GP classification tasks. This work remediates these issues by basing our model on sigmoidal Gaussian Cox process (SGCP)~\citep{adams2009tractable}, using a scaled sigmoid function as link function in point process tasks, and the logistic regression model in classification tasks. The reason we choose sigmoid as link function in both types of tasks is we can exploit the data augmentation technique~\citep{polson2013bayesian,donner2018efficient} to construct a mean-field approximation that has closed-form iterative updates. As shown later, the proposed mean-field approximation exhibits superior efficiency and fast convergence. 

Specifically, we make the following contributions. \textbf{(1)} From a modeling perspective, we establish a MOGP based \emph{heterogeneous multi-task Gaussian Cox processes} (HMGCP) model that provides an extension of the homogeneous version to account for multiple heterogeneous correlated tasks. \textbf{(2)} From an inference perspective, we 
adopt the data augmentation technique to derive \emph{an efficient mean-field approximation with analytical expressions}. As far as we know, this work should be the first attempt to use data augmentation in the MOGP setting. 
\textbf{(3)} In experiments, we provide evidence of the benefits of modeling heterogeneous correlated tasks and the predominant efficiency and convergence of our inference method. 

\section{Related Work}
\paragraph{Multi-Output Gaussian Processes}
Multi-output Gaussian processes~\citep{alvarezRL12kernel} extend the single-output Gaussian process to model vector-valued functions, providing a powerful Bayesian tool for multi-task learning as it accounts for the correlation between multiple outputs. \citet{bonillaCW07} has shown that if multiple outputs are correlated, exploiting such correlation can provide insightful information about each output and better predictions in the case of sparse/missing data. More importantly, as a Bayesian nonparametric approach, it offers higher flexibility over parametric alternatives and a natural mechanism for uncertainty quantification. 
To define a MOGP, we need to define a suitable cross-covariance function that accounts for the correlation between multiple outputs, which leads to a valid covariance function for the joint GP~\citep{alvarez2019non}. The two common ways to define cross-covariance functions are linear model of coregionalization (LMC)~\citep{journel1976mining} and process convolution~\citep{ver1998constructing}. In this work, we focus on the LMC approach.



\paragraph{Multi-Task Cox Processes}
Extensive works have been accumulated on the single-task Gaussian Cox process~\citep{moller1998log,diggle2013spatial}. 
Recently, many works tried to extend the single-task Cox process to the multi-task setup to introduce correlation between tasks. 
For example, \citet{lian2015multitask} proposed a multi-task Cox process model that leverages information from all tasks via a hierarchical GP. In a different way, \citet{aglietti2019efficient} and \citet{jahani2021multioutput} adopted the MOGP based on LMC and process convolution respectively to model the intensity functions of multiple Cox processes, which facilitates sharing of information and allows for flexible event occurrence rate. All these works exclusively focus on homogeneous multi-task Cox processes. On the contrary, we extend to the heterogeneous scenarios to enable transfer of knowledge between Cox process, regression and classification tasks. 

\paragraph{Data Augmentation}
In GP regression, the conjugacy between likelihood and prior makes the posterior computing easy and closed-form. However, in GP classification and point process, such conjugacy no longer holds and one may resort to variational inference to approximate the true posterior. Most generic non-conjugate variational inference, assuming a Gaussian variational posterior to make the ELBO tractable, exhibits low efficiency due to computing of expectations~\citep{dezfouli2015scalable}. Recently, another inference method based on data augmentation\footnote{The notion of data augmentation in statistics is different from that in deep learning.} has been established for GP classification~\citep{polson2013bayesian,wenzel2019efficient} and point process~\citep{donner2018efficient,zhou2020auxiliary,zhou2021efficient,zhou2022efficient}. The core idea is to augment likelihood by auxiliary latent variables to convert the non-conjugate problem to a conditionally conjugate one, thus making inference easy~\citep{li2014bayesian}. Here, such an idea is extended to the MOGP modulated multi-task framework. 

\section{Problem Formulation}
Traditionally, existing works have considered the homogeneous multi-task Cox processes learning where all tasks are Cox processes~\citep{aglietti2019efficient,jahani2021multioutput}.
The homogeneous model is not applicable to the more general heterogeneous scenario which includes various types of tasks except Cox processes. 
In this work, we are interested in the more general heterogeneous scenario where correlated tasks are a mix of supervised (regression and classification) and unsupervised tasks (Cox processes). 
Let us consider a problem setting where we have data from $I$ tasks, among which $I_r$ tasks are regression problems with dataset $\mathcal{D}_r=\{\{(\mathbf{x}_{i,n}^r,y_{i,n}^r)\}_{n=1}^{N_i^r}\}_{i=1}^{I_r}$, $I_c$ tasks are classification problems with dataset $\mathcal{D}_c=\{\{(\mathbf{x}_{i,n}^c,y_{i,n}^c)\}_{n=1}^{N_i^c}\}_{i=1}^{I_c}$ and $I_p$ tasks are point process problems with dataset $\mathcal{D}_p=\{\{(\mathbf{x}_{i,n}^p)\}_{n=1}^{N_i^p}\}_{i=1}^{I_p}$. $\mathbf{x}\in\mathcal{X}\subset\mathbb{R}^D$ is the $D$-dimensional input; $y\in\mathbb{R}$ is the output in regression tasks and $\{-1,1\}$ in classification tasks\footnote{We focus on binary classification here. Extension to multi-class classification is discussed in \cref{app.multiclass}.}.
Point process tasks are unsupervised learning problems so they only include $\mathbf{x}$. 
Throughout the paper, we use index $r,c,p$ to indicate regression, classification and point process tasks, respectively. 

\subsection{Heterogeneous Likelihood}
In order to use GP to represent the likelihood parameters in three types of tasks, we need to design the appropriate transformation to map the GP output to the domain of specific parameters. For regression tasks, following tradition, we use Gaussian distribution as likelihood, where the mean is modeled as a GP function and the variance is treated as a hyperparameter. For binary classification tasks, we use Bernoulli distribution~\citep{uspensky1937introduction} as likelihood whose parameter is modeled by the sigmoid transformation of a GP function, mapping $\mathbb{R}\to[0,1]$, which is also called logistic regression. For Cox process tasks, although many existing works focus on LGCP, our work adopts the SGCP instead, i.e., the intensity of $i$-th Cox process is assumed to be $\lambda_i(\mathbf{x})=\bar{\lambda}_i s(g_i(\mathbf{x}))$ where a task-specific GP function $g_{i}$ is passed through a sigmoid function $s(\cdot)$ and then scaled by an upper-bound $\bar{\lambda}_{i}$. The reason we choose the sigmoid link function in both classification and point process tasks is that we can exploit the data augmentation to make inference easy and fast. Specifically, three types of likelihoods are: 
\begin{subequations}
\label{eq1}
\begin{gather}
\label{eq1a}
p(\mathbf{y}^{r}\mid\{g_{i}^r\}_{i=1}^{I_r})=\prod_{i=1}^{I_r}\prod_{n=1}^{N_{i}^r}\mathcal{N}(y^r_{i,n}\mid g^r_{i,n},\sigma_i^2),\\
\label{eq1b}
p(\mathbf{y}^{c}\mid\{g_{i}^c\}_{i=1}^{I_c})=\prod_{i=1}^{I_c}\prod_{n=1}^{N^c_{i}}s(y^c_{i,n}g^c_{i,n}),\\
\label{eq1c}
p(\mathbf{x}^p\mid\{\bar{\lambda}_i,g^p_i\}_{i=1}^{I_p})=\prod_{i=1}^{I_p}\prod_{n=1}^{N^p_i}\bar{\lambda}_i s(g^p_{i,n})\exp\left(-\int_{\mathcal{X}}\bar{\lambda}_i
s(g^p_i(\mathbf{x}))d\mathbf{x}\right),
\end{gather}
\end{subequations}
where $g_{i}$ is the task-specific GP function and we call it latent function~\citep{rasmussen2003gaussian} afterwards; $g_{i}^r$, $g_{i}^c$, $g^p_i$ are the corresponding $i$-th output of the regression, classification and point process tasks, respectively; $g^{\cdot}_{i,n}$ indicates $g^{\cdot}_i(\mathbf{x}^{\cdot}_{i,n})$. \Cref{eq1a} is the likelihood for regression; \cref{eq1b} is the likelihood for binary classification; \cref{eq1c} is the likelihood for point process~\citep{daley2003introduction}. 

\subsection{MOGP Prior}
Instead of modeling each $g_i$ independently, we apply the MOGP prior on $g$'s to introduce correlation between multiple tasks in order to improve the generalization capability of our model especially when data is sparse or missing. In this work, we use the LMC~\citep{journel1976mining} approach to define the cross-covariance function. Specifically, we assume each latent function $g_i$ is a linear combination of $Q$ basis functions which are drawn from $Q$ independent zero-mean GP prior, i.e., $\{f_q\sim\mathcal{GP}(0,k_q)\}_{q=1}^Q$ where $k_q$ is a covariance function. Each latent function can be written as $g_i=\sum_{q=1}^Qw_{i,q}f_q$ where $w_{i,q}\in\mathbb{R}$ is the mixing weight capturing the contribution of $q$-th basis function to $i$-th latent function. It is easy to see that the mean of $g_i$ is zero and the cross-covariance $k_{g_i,g_j}(\mathbf{x},\mathbf{x}')=\text{cov}[g_i(\mathbf{x}),g_j(\mathbf{x}')]=\sum_{q=1}^Qw_{i,q}w_{j,q}k_q(\mathbf{x},\mathbf{x}')$. If we define $\mathbf{g}_i$ to be the vector of latent function values on the inputs of $i$-th task, we have the following MOGP prior: $\mathbf{g}\sim\mathcal{N}(\mathbf{0},\mathbf{K})$, where $\mathbf{g}=[\mathbf{g}_{1}^\top,\ldots,\mathbf{g}_{I}^\top]^\top$, $I=I_r+I_c+I_p$, $\mathbf{K}$ is a block-wise matrix with blocks given by $\{\mathbf{K}_{\mathbf{g}_{i},\mathbf{g}_{j}}\}_{i=1,j=1}^{I,I}$ whose entries are $k_{g_i,g_j}(\mathbf{x},\mathbf{x}')$. $\mathbf{x}$ and $\mathbf{x}'$ are the inputs of $i$-th and $j$-th tasks, respectively. It is worth noting that each task can have a different set of inputs, but when all tasks have the same set of inputs, e.g., $\mathbf{X}$, the computing of $\mathbf{K}$ can be simplified as the sum of Kronecker products $\mathbf{K}=\sum_{q=1}^Q\mathbf{w}_q\mathbf{w}_q^\top\otimes\mathbf{K}_q$ where $\mathbf{w}_q=[w_{1,q},\ldots, w_{I,q}]^\top$, $\mathbf{K}_q$ is the square matrix of $k_q(\mathbf{x},\mathbf{x}')$ with $\mathbf{x},\mathbf{x}'\in\mathbf{X}$~\citep{moreno2018heterogeneous}. This property cooperates well with the inducing inputs formalism which is discussed later. 

\section{Inference}
According to Bayes' theorem, the posterior of latent functions and intensity upper-bounds can be computed as:
\begin{equation*}
\begin{aligned}
&p(g,\bar{\bm{\lambda}}\mid\mathbf{y}^{r},\mathbf{y}^{c},\mathbf{x}^p)\propto\\
&\underbrace{p(\mathbf{y}^{r}\mid\{g_{i}^r\}_{i=1}^{I_r})}_{\text{regression}}\underbrace{p(\mathbf{y}^{c}\mid\{g_{i}^c\}_{i=1}^{I_c})}_{\text{classification}}\underbrace{p(\mathbf{x}^p\mid\{\bar{\lambda}_i,g^p_i\}_{i=1}^{I_p})}_{\text{Cox process}}\underbrace{p(g)}_{\text{MOGP}}p(\bar{\bm{\lambda}}), 
\end{aligned}
\label{eq3}
\end{equation*}
where $g=[g_1,\ldots,g_I]^\top$, $\bar{\bm{\lambda}}=[\bar{\lambda}_1,\ldots,\bar{\lambda}_{I_p}]^\top$, $p(g)$ is the infinite-dimensional version of MOGP, $p(\bar{\bm{\lambda}})\propto\prod_{i=1}^{I_p}\frac{1}{\bar{\lambda}_i}$ is the improper prior. The likelihood of regression is conjugate to the prior. However, such conjugacy is no longer valid for classification and Cox process tasks, so the posterior has no closed-form solution. 

To address the non-conjugate issue for classification or Cox process, many works applied the variational inference that assumed a Gaussian variational distribution to render the ELBO tractable~\citep{dezfouli2015scalable,hensman2015scalable,aglietti2019efficient,jahani2021multioutput}. However, such generic variational inference exhibits low efficiency due to computing of expectations in ELBO~\citep{wenzel2019efficient}. In this work, borrowing the idea of data augmentation, we augment P\'{o}lya-Gamma latent variables~\citep{polson2013bayesian} and marked Poisson latent processes~\citep{donner2018efficient} into the likelihood of classification and Cox process. Finally, the augmented likelihood is conditionally conjugate to the MOGP prior. Based on the augmented model, we derive a mean-field approximation with closed-form iterative updates to provide an approximate posterior. The proofs of all relevant formulas below are provided in the appendix. 

\subsection{Augmentation for Classification Tasks}
\citet{polson2013bayesian} proposed a novel P\'{o}lya-Gamma augmentation strategy for Bayesian logistic regression. The core idea is that the binomial likelihood parametrized by log odds can be represented as a mixture of Gaussians w.r.t. a P\'{o}lya-Gamma distribution.

If $\omega\sim p_{\text{PG}}(\omega\mid b,0)$ denotes the P\'{o}lya-Gamma random variable with $\omega\in \mathbb{R}^+$ and $b>0$, the following integral identity holds for $a\in\mathbb{R}$: 
\begin{equation*}
\frac{(e^{z})^a}{(1+e^{z})^b}=2^{-b}e^{(a-b/2)z}\int_0^\infty e^{-z^2\omega/2}p_{\text{PG}}(\omega\mid b,0)d\omega. 
\end{equation*}
In this work, we do not need to know the exact form of the P\'{o}lya-Gamma distribution, but only its first moment. Setting $a=b=1$ yields the factorization of sigmoid function: 
\begin{equation}
s(z)=\frac{e^z}{1+e^z}=\int_0^{\infty} e^{h(\omega,z)}p_{\text{PG}}(\omega\mid1,0)d\omega,
\label{eq4}
\end{equation}
where $h(\omega,z)=z/2-z^2\omega/2-\log2$. Substituting \cref{eq4} into the classification likelihood in \cref{eq1b}, we obtain the augmented classification likelihood which has the elegant conditionally conjugate property. 
After augmenting P\'{o}lya-Gamma random variables, the logistic regression likelihood in \cref{eq1b} is augmented to be: 
\begin{equation}
\begin{aligned}
&p(\mathbf{y}^c,\bm{\omega}^c\mid\{g^c_i\}_{i=1}^{I_c})=\prod_{i=1}^{I_c}\prod_{n=1}^{N^c_i}e^{h(\omega^c_{i,n},y^c_{i,n}g^c_{i,n})}p_{\text{PG}}(\omega^c_{i,n}\mid1,0),
\end{aligned}
\label{eq5}
\end{equation}
where $\omega^c_{i,n}$ is the P\'{o}lya-Gamma latent variable on the $n$-th observed sample in the $i$-th classification task, $\bm{\omega}^c_i=[\omega^c_{i,1},\ldots,\omega^c_{i,N_i^c}]^\top$, $\bm{\omega}^c=[{\bm{\omega}^c_1}^\top,\ldots,{\bm{\omega}^c_{I_c}}^\top]^\top$. The derivation is provided in \cref{app.corollary1}. The augmented classification likelihood in \cref{eq5} is conditionally conjugate to the MOGP prior.

\subsection{Augmentation for Cox Process Tasks}
The augmentation for Cox process is more challenging than classification because the Cox process likelihood depends 
not only on the latent function values on observed samples but also 
on the whole latent function due to the exponential integral term. Borrowing the idea from~\citet{donner2018efficient}, in addition to augmenting P\'{o}lya-Gamma latent variables on observed samples as in classification tasks, we also augment a marked Poisson latent process to linearize the exponential integral term. 

Define a marked Poisson process $\Pi=\{(\mathbf{x}_r,\omega_r)\}\sim p(\Pi\mid\bar{\lambda}p_{\text{PG}}(\omega\mid1,0))$ where $\mathbf{x}_r$ is the location of $r$-th point, the P\'{o}lya-Gamma latent variable $\omega_r$ denotes the independent mark at each point $\mathbf{x}_r$, $p(\Pi\mid\bar{\lambda}p_{\text{PG}}(\omega\mid1,0))$ denotes the probability measure of $\Pi$ with intensity $\Lambda(\mathbf{x},\omega)=\bar{\lambda}p_{\text{PG}}(\omega\mid1,0)$. Given the marked Poisson process defined above, the following identity holds: 
\begin{equation}
\exp{\left(-\int_\mathcal{X}\bar{\lambda}s(g(\mathbf{x}))d\mathbf{x}\right)}=\mathbb{E}_{p_\Lambda}\prod_{(\omega,\mathbf{x})\in\Pi}e^{h(\omega,-g(\mathbf{x}))}, 
\label{eq6}
\end{equation}
where $p_\Lambda$ indicates $p(\Pi\mid\Lambda(\mathbf{x},\omega)=\bar{\lambda}p_{\text{PG}}(\omega\mid1,0))$.
Substituting \cref{eq4,eq6} into 
the Cox process likelihood in \cref{eq1c}, we obtain the augmented Cox process likelihood which has the conditionally conjugate property. 
After augmenting the P\'{o}lya-Gamma latent variables on observed samples and the marked Poisson latent process, the Cox process likelihood in \cref{eq1c} is augmented to be: 
\begin{equation}
\begin{aligned}
&p(\mathbf{x}^p,\bm{\omega}^p,\Pi\mid\bar{\bm{\lambda}}, \{g^p_i\}_{i=1}^{I_p})=\\
&\prod_{i=1}^{I_p}\prod_{n=1}^{N_i^p}\Lambda_{i}(\mathbf{x}^p_{i,n},\omega^p_{i,n})e^{h(\omega^p_{i,n},g^p_{i,n})}p_{\Lambda_i}(\Pi_i\mid\bar{\lambda}_i)\prod_{(\omega,\mathbf{x})\in\Pi_i}e^{h(\omega,-g^p_i(\mathbf{x}))}, 
\end{aligned}
\label{eq7}
\end{equation}
where $\omega^p_{i,n}$ is the P\'{o}lya-Gamma latent variable on $n$-th observed sample in the $i$-th Cox process task, $\bm{\omega}^p_i=[\omega^p_{i,1},\ldots,\omega^p_{i,N_i^p}]^\top$, $\bm{\omega}^p=[{\bm{\omega}^p_1}^\top,\ldots,{\bm{\omega}^p_{I_p}}^\top]^\top$, $\Lambda_{i}(\mathbf{x},\omega)=\bar{\lambda}_ip_{\text{PG}}(\omega\mid1,0)$, $\Pi=\{\Pi_i\}_{i=1}^{I_p}$. The derivation is provided in \cref{app.corollary2}. The augmented Cox process likelihood in \cref{eq7} is conditionally conjugate to the MOGP prior.

\subsection{Mean-Field Approximation}
Based on the augmented likelihoods for classification and Cox process in \cref{eq5,eq7}, we obtain the augmented joint distribution for all variables:
\begin{equation}
\begin{aligned}
&p(\mathbf{y}^{r},\mathbf{y}^{c},\mathbf{x}^p,\bm{\omega}^c,\bm{\omega}^p,\Pi,g,\bar{\bm{\lambda}})=\\
&\underbrace{p(\mathbf{y}^{r}\mid\{g_{i}^r\}_{i=1}^{I_r})}_{\text{regression}}\underbrace{p(\mathbf{y}^c,\bm{\omega}^c\mid\{g^c_i\}_{i=1}^{I_c})}_{\text{augmented classification}}\underbrace{p(\mathbf{x}^p,\bm{\omega}^p,\Pi\mid\bar{\bm{\lambda}}, \{g^p_i\}_{i=1}^{I_p})}_{\text{augmented Cox process}}\underbrace{p(g)}_{\text{MOGP}}p(\bar{\bm{\lambda}}). 
\end{aligned}
\label{eq8}
\end{equation}
Finally, our efforts are rewarded: after data augmentation, the model likelihood is conditionally conjugate to the prior and a simple Gibbs sampler can be derived to sample from the \emph{exact posterior} by drawing a sample from each conditional distribution alternately. 
The samples of latent functions and intensity upper-bounds will be from the true posterior asymptotically. 
However, the sampling approach has a prohibitive computational cost and does not scale to large datasets. The comparison of efficiency between Gibbs sampler and variational inference is outside of the scope of this paper. 
Here we adopt the augmented model to derive an efficient mean-field approximation, which has closed-form iterative updates. 

Following the standard derivation of mean-field approximation, we assume the posterior $p(\bm{\omega}^c,\bm{\omega}^p,\Pi,g,\bar{\bm{\lambda}}\mid\mathbf{y}^{r},\mathbf{y}^{c},\mathbf{x}^p)$ is approximated by a variational posterior: 
\begin{equation*}
q(\bm{\omega}^c,\bm{\omega}^p,\Pi,g,\bar{\bm{\lambda}})=q_1(\bm{\omega}^c,\bm{\omega}^p,\Pi)q_2(g,\bar{\bm{\lambda}}). 
\end{equation*}
The independence of two sets of variables is the only assumption of the variational posterior. To minimize the Kullback–Leibler (KL) divergence between $q$ and $p$, it can be proved that the optimal distribution of each factor is the expectation of the logarithm of \cref{eq8} taken over variables in the other factor~\citep{bishop2006pattern,blei2017variational}: 
\begin{equation}
\begin{aligned}
q_1^*(\bm{\omega}^c,\bm{\omega}^p,\Pi)&\propto e^{{\mathbb{E}_{q_2}[\log p(\mathbf{y}^{r},\mathbf{y}^{c},\mathbf{x}^p,\bm{\omega}^c,\bm{\omega}^p,\Pi,g,\bar{\bm{\lambda}})]}},\\
q_2^*(g,\bar{\bm{\lambda}})&\propto e^{{\mathbb{E}_{q_1}[\log p(\mathbf{y}^{r},\mathbf{y}^{c},\mathbf{x}^p,\bm{\omega}^c,\bm{\omega}^p,\Pi,g,\bar{\bm{\lambda}})]}}.
\end{aligned}
\label{eq9}
\end{equation}

A prominent weakness of GP is that it suffers from a cubic complexity w.r.t. the number of samples. 
In multi-task scenario, although the samples in a single task can be few, the total number of samples in all tasks can be large. 
To make our mean-field approximation scalable, we employ the inducing points formalism~\citep{alvarez2008sparse,titsias2009variational}. We denote $M$ inducing inputs $[\mathbf{x}_1\,\ldots,\mathbf{x}_M]^\top$ on the domain $\mathcal{X}$ for each task. The function values of basis function $f_q$ at these inducing inputs are defined as $\mathbf{f}_{q,\mathbf{x}_m}$. Then we can obtain the $i$-th task latent function $g_i$ at these inducing inputs $\mathbf{g}_{\mathbf{x}_m}^i=\sum_{q=1}^Qw_{i,q}\mathbf{f}_{q,\mathbf{x}_m}$\footnote{For the compactness of notation, the task index $i$ is sometimes moved from subscript to superscript, which does not cause confusion because we use $i$ consistently.}. If we define $\mathbf{g}_{\mathbf{x}_m}=[\mathbf{g}^{1\top}_{\mathbf{x}_m},\ldots,\mathbf{g}^{I\top}_{\mathbf{x}_m}]^\top$, $\mathbf{g}_{\mathbf{x}_m}\sim\mathcal{N}(\mathbf{0},\mathbf{K}_{\mathbf{x}_m \mathbf{x}_m})$ where $\mathbf{K}_{\mathbf{x}_m\mathbf{x}_m}$ is the MOGP covariance on $\mathbf{x}_m$ for all tasks and $\mathbf{g}_{\mathbf{x}_m}^i\sim\mathcal{N}(\mathbf{0},\mathbf{K}_{\mathbf{x}_m \mathbf{x}_m}^i)$ where $\mathbf{K}_{\mathbf{x}_m\mathbf{x}_m}^i$ is $i$-th diagonal block of $\mathbf{K}_{\mathbf{x}_m\mathbf{x}_m}$. Given $\mathbf{g}_{\mathbf{x}_m}^i$, we assume $p(g_i(\mathbf{x})\mid\mathbf{g}_{\mathbf{x}_m}^i)=\mathcal{N}(\mathbf{k}_{\mathbf{x}_m \mathbf{x}}^{i\top}\mathbf{K}_{\mathbf{x}_m \mathbf{x}_m}^{i^{-1}}\mathbf{g}_{\mathbf{x}_m}^i, k_{\mathbf{x}\mathbf{x}}^{i}-\mathbf{k}_{\mathbf{x}_m \mathbf{x}}^{i\top}\mathbf{K}_{\mathbf{x}_m \mathbf{x}_m}^{i^{-1}}\mathbf{k}_{\mathbf{x}_m \mathbf{x}}^{i})$ where $\mathbf{k}_{\mathbf{x}_m \mathbf{x}}^{i}$ is the kernel w.r.t. inducing points and the predictive point, $k_{\mathbf{x}\mathbf{x}}^{i}$ is the kernel w.r.t. the predictive point for $i$-th task. 

After substituting \cref{eq8} into \cref{eq9} and introducing the inducing points, we can obtain the optimal variational distribution of each factor in the following closed-form expressions (derivation provided in \cref{app.mean-field}): 
\paragraph{The Optimal Density of P\'{o}lya-Gamma Latent Variables}
The optimal variational posteriors of $\bm{\omega}^c$ and $\bm{\omega}^p$ are: 
\begin{equation}
q_1(\bm{\omega}^c)=\prod_{i=1}^{I_c}\prod_{n=1}^{N_i^c}p_{\text{PG}}(\omega^c_{i,n}\mid1,\tilde{g}^c_{i,n}), 
\label{eq10a}
\end{equation}
\begin{equation}
q_1(\bm{\omega}^p)=\prod_{i=1}^{I_p}\prod_{n=1}^{N_i^p}p_{\text{PG}}(\omega^p_{i,n}\mid1,\tilde{g}^p_{i,n}), 
\label{eq10b}
\end{equation}
where $\tilde{g}^\cdot_{i,n}=\sqrt{\mathbb{E}[{g^\cdot_{i,n}}^2]}$. 

\paragraph{The Optimal Intensity of Marked Poisson Processes}
The optimal variational posterior intensity of $\Pi=\{\Pi_i\}_{i=1}^{I_p}$ is: 
\begin{equation}
\begin{aligned}
\Lambda_i^1(\mathbf{x},\omega)=\bar{\lambda}_i^1 s(-\tilde{g}^p_i(\mathbf{x}))p_{\text{PG}}(\omega\mid 1,\tilde{g}^p_i(\mathbf{x}))e^{(\tilde{g}^p_i(\mathbf{x})-\bar{g}^p_i(\mathbf{x}))/2},
\end{aligned}
\label{eq10c}
\end{equation}
where $\bar{\lambda}^1_i=e^{\mathbb{E}[\log\bar{\lambda}_i]}$, $\tilde{g}_i^p(\mathbf{x})=\sqrt{\mathbb{E}[{g_i^p(\mathbf{x})}^2]}$ and $\bar{g}_i^p(\mathbf{x})=\mathbb{E}[g_i^p(\mathbf{x})]$. 

\paragraph{The Optimal Density of Intensity Upper-bounds}
The optimal variational posterior of $\bar{\bm{\lambda}}$ is: 
\begin{equation}
\begin{aligned}
q_2(\bar{\bm{\lambda}})=\prod_{i=1}^{I_p}p_{\text{Ga}}(\bar{\lambda}_i\mid N_i^p+R_i,\lvert\mathcal{X}\rvert),
\end{aligned}
\label{eq10d}
\end{equation}
where $p_{\text{Ga}}$ is Gamma density, $R_i=\int_\mathcal{X}\int_0^\infty\Lambda_i^1(\mathbf{x},\omega)d\omega d\mathbf{x}$, $\lvert\mathcal{X}\rvert$ is the domain size. 

\paragraph{The Optimal Density of Latent Functions}
The optimal variational posterior of $\mathbf{g}_{\mathbf{x}_m}$ is: 
\begin{equation}
\begin{gathered}
q_2(\mathbf{g}_{\mathbf{x}_m})=\mathcal{N}(\mathbf{g}_{\mathbf{x}_m}\mid\mathbf{m}_{\mathbf{x}_m},\mathbf{\Sigma}_{\mathbf{x}_m}),
\end{gathered}
\label{eq10e}
\end{equation}
where $\mathbf{g}_{\mathbf{x}_m}=[\mathbf{g}_{\mathbf{x}_m}^{r\top},\mathbf{g}_{\mathbf{x}_m}^{c\top},\mathbf{g}_{\mathbf{x}_m}^{p\top}]^\top$, $\mathbf{g}_{\mathbf{x}_m}^{\cdot}=[\mathbf{g}_{1,\mathbf{x}_m}^{\cdot\top},\ldots,\mathbf{g}_{I_\cdot,\mathbf{x}_m}^{\cdot\top}]^\top$ and 
\begin{equation*}
\mathbf{\Sigma}_{\mathbf{x}_m}=\left[\text{diag}\left(\mathbf{H}_{\mathbf{x}_m}^r,\mathbf{H}_{\mathbf{x}_m}^c,\mathbf{H}_{\mathbf{x}_m}^p\right)+\mathbf{K}_{\mathbf{x}_m \mathbf{x}_m}^{-1}\right]^{-1},\mathbf{m}_{\mathbf{x}_m}=\mathbf{\Sigma}_{\mathbf{x}_m}[\mathbf{v}_{\mathbf{x}_m}^{r\top},\mathbf{v}_{\mathbf{x}_m}^{c\top},\mathbf{v}_{\mathbf{x}_m}^{p\top}]^\top,
\end{equation*}
where $\mathbf{H}_{\mathbf{x}_m}^\cdot=\text{diag}(\mathbf{H}_{1,\mathbf{x}_m}^\cdot,\ldots,\mathbf{H}_{I_\cdot,\mathbf{x}_m}^\cdot)$, $\mathbf{v}_{\mathbf{x}_m}^\cdot=[\mathbf{v}_{1,\mathbf{x}_m}^{\cdot\top},\ldots,\mathbf{v}_{I_\cdot,\mathbf{x}_m}^{\cdot\top}]^\top$ and 
\begin{equation*}
\begin{gathered}
\mathbf{H}_{i,\mathbf{x}_m}^r=\mathbf{K}_{\mathbf{x}_m \mathbf{x}_m}^{r,i^{-1}}\mathbf{K}_{\mathbf{x}_m \mathbf{x}_n}^{r,i}\mathbf{D}^r_i\mathbf{K}_{\mathbf{x}_m \mathbf{x}_n}^{r,i\top}\mathbf{K}_{\mathbf{x}_m \mathbf{x}_m}^{r,i^{-1}},
\mathbf{v}_{i,\mathbf{x}_m}^{r}=\mathbf{K}_{\mathbf{x}_m \mathbf{x}_m}^{r,i^{-1}}\mathbf{K}_{\mathbf{x}_m \mathbf{x}_n}^{r,i}\frac{\mathbf{y}^r_i}{\sigma_i^2},\\
\mathbf{H}_{i,\mathbf{x}_m}^c=\mathbf{K}_{\mathbf{x}_m \mathbf{x}_m}^{c,i^{-1}}\mathbf{K}_{\mathbf{x}_m \mathbf{x}_n}^{c,i}\mathbf{D}^c_i\mathbf{K}_{\mathbf{x}_m \mathbf{x}_n}^{c,i\top}\mathbf{K}_{\mathbf{x}_m \mathbf{x}_m}^{c,i^{-1}},
\mathbf{v}_{i,\mathbf{x}_m}^{c}=\mathbf{K}_{\mathbf{x}_m \mathbf{x}_m}^{c,i^{-1}}\mathbf{K}_{\mathbf{x}_m \mathbf{x}_n}^{c,i}\frac{\mathbf{y}^c_i}{2},\\
\mathbf{H}_{i,\mathbf{x}_m}^p=\mathbf{K}_{\mathbf{x}_m \mathbf{x}_m}^{p,i^{-1}}\int_{\mathcal{X}}A_i(\mathbf{x})\mathbf{k}_{\mathbf{x}_m \mathbf{x}}^{p,i}\mathbf{k}_{\mathbf{x}_m \mathbf{x}}^{p,i\top}d\mathbf{x}\mathbf{K}_{\mathbf{x}_m \mathbf{x}_m}^{p,i^{-1}},\\
\mathbf{v}_{i,\mathbf{x}_m}^{p}=\mathbf{K}_{\mathbf{x}_m \mathbf{x}_m}^{p,i^{-1}}\int_{\mathcal{X}}B_i(\mathbf{x})\mathbf{k}_{\mathbf{x}_m \mathbf{x}}^{p,i}d\mathbf{x},
\end{gathered}
\end{equation*}
where $\mathbf{D}^r_i=\text{diag}(1/\sigma_i^2)$, $\mathbf{D}^c_i=\text{diag}(\mathbb{E}[\bm{\omega}^{c}_i])$ and 
\begin{equation*}
\begin{gathered}
    A_i(\mathbf{x})=\sum_{n=1}^{N^p_i}\mathbb{E}[\omega_{i,n}^p]\delta(\mathbf{x}-\mathbf{x}^p_{i,n})+\int_0^\infty\omega\Lambda_i^1(\mathbf{x},\omega)d\omega,\\
    B_i(\mathbf{x})=\frac{1}{2}\sum_{n=1}^{N^p_i}\delta(\mathbf{x}-\mathbf{x}^p_{i,n})-\frac{1}{2}\int_0^\infty\Lambda_i^1(\mathbf{x},\omega)d\omega. 
\end{gathered}
\end{equation*}

\paragraph{Predictive Distribution}
The posterior distribution of the task-specific latent function $g_i$ at a predictive point $\mathbf{x}$ is approximated by 
\begin{equation*}
    q(g_i(\mathbf{x}))=\int p(g_i(\mathbf{x})\mid\mathbf{g}_{\mathbf{x}_m}^i)q(\mathbf{g}_{\mathbf{x}_m}^i)d\mathbf{g}_{\mathbf{x}_m}^i=\mathcal{N}(g_i(\mathbf{x})\mid\mu,\sigma^2),
\end{equation*}
where $\mu=\mathbf{k}_{\mathbf{x}_m \mathbf{x}}^{i\top}\mathbf{K}_{\mathbf{x}_m \mathbf{x}_m}^{i^{-1}}\mathbf{m}_{\mathbf{x}_m}^i$, $\sigma^2=k_{\mathbf{x}\mathbf{x}}^{i}-\mathbf{k}_{\mathbf{x}_m \mathbf{x}}^{i\top}\mathbf{K}_{\mathbf{x}_m \mathbf{x}_m}^{i^{-1}}\mathbf{k}_{\mathbf{x}_m \mathbf{x}}^{i}+\mathbf{k}_{\mathbf{x}_m \mathbf{x}}^{i\top}\mathbf{K}_{\mathbf{x}_m \mathbf{x}_m}^{i^{-1}}\mathbf{\Sigma}_{\mathbf{x}_m}^i\mathbf{K}_{\mathbf{x}_m\mathbf{x}_m}^{i^{-1}}\mathbf{k}_{\mathbf{x}_m \mathbf{x}}^{i}$. Therefore, $\tilde{g}_i(\mathbf{x})=\sqrt{\mu^2+\sigma^2}$, $\bar{g}_i(\mathbf{x})=\mu$, $\mathbb{E}[\omega]=\frac{b}{2c}\tanh\frac{c}{2}$ for $p_{\text{PG}}(\omega\mid b,c)$~\citep{polson2013bayesian}, $\mathbb{E}[\log\bar{\lambda}_i]=\psi(N_i^p+R_i)-\log(\lvert\mathcal{X}\rvert)$ where $\psi(\cdot)$ is digamma function. The intractable integral over $\mathcal{X}$ is solved by numerical quadrature. Updating the variational posterior of each factor alternately by \cref{eq10a,eq10b,eq10c,eq10d,eq10e}, we obtain approximate posteriors of $\bar{\bm{\lambda}}$ and $\mathbf{g}_{\mathbf{x}_m}$. 

\paragraph{Hyperparameters and Computation Complexity}
The model hyperparameter $\bm{\Theta}$ comprises the kernel hyperparameters $\{\bm{\theta}_q\}_{q=1}^Q$ associated to the covariance functions $\{k_q\}_{q=1}^Q$, the mixing weights $\{\mathbf{w}_q\}_{q=1}^Q$, the inducing inputs $\{\mathbf{x}_m\}_{m=1}^M$ and the noise variance $\{\sigma_i^2\}_{i=1}^{I_r}$ in regression tasks. In this work, the inducing points are uniformly located on the domain, which means the kernel matrix has Toeplitz structure~\citep{cunningham2008fast} and this can lead to more efficient matrix inversion.
In the implementation, we do not apply this method and instead use the naive matrix inversion.
$\{\bm{\theta}_q\}_{q=1}^Q$, $\{\mathbf{w}_q\}_{q=1}^Q$ and $\{\sigma_i^2\}_{i=1}^{I_r}$ are optimized by maximizing the marginal likelihood, which is also called the empirical Bayes. Due to the intractability of marginal likelihood, we adopt an approximate approach: maximize the ELBO as a function of hyperparameters by alternating between updating variational parameters and hyperparameters. In the following, we derive the ELBO: 
\begin{equation*}
\begin{aligned}
&\log p(\mathbf{y}^{r},\mathbf{y}^{c},\mathbf{x}^p)\geq\\
&\mathbb{E}_q[\log p(\mathbf{y}^{r},\mathbf{y}^{c},\mathbf{x}^p\mid\bm{\omega}^c,\bm{\omega}^p,\Pi,g,\bar{\bm{\lambda}})]-\text{KL}(q(\bm{\omega}^c,\bm{\omega}^p,\Pi,g,\bar{\bm{\lambda}})\Vert p(\bm{\omega}^c,\bm{\omega}^p,\Pi,g,\bar{\bm{\lambda}}))\\
&=\mathbb{E}_q[\log p(\mathbf{y}^{r}\mid \{g_{i}^r\}_{i=1}^{I_r})]+\mathbb{E}_q[\log p(\mathbf{y}^{c}\mid\bm{\omega}^c,\{g_{i}^c\}_{i=1}^{I_c})]\\
&+\mathbb{E}_q[\log p(\mathbf{x}^p\mid\bm{\omega}^p,\Pi,\{g_{i}^p\}_{i=1}^{I_p},\bar{\bm{\lambda}})]-\text{KL}(q(g)\Vert p(g))\\
&-\text{KL}(q(\bm{\omega}^c,\bm{\omega}^p,\Pi,\bar{\bm{\lambda}})\Vert p(\bm{\omega}^c,\bm{\omega}^p,\Pi,\bar{\bm{\lambda}})),
\end{aligned}
\label{app.eq15}
\end{equation*}
where we omit the conditioning on hyperparameters. It is straightforward to see that, given variational posteriors, only the first term includes the noise variance $\{\sigma_i^2\}_{i=1}^{I_r}$ and only the fourth term includes the kernel hyperparameters $\{\bm{\theta}_q\}_{q=1}^Q$ and the mixing weights $\{\mathbf{w}_q\}_{q=1}^Q$. All other terms are constant w.r.t. hyperparameters. After introducing the inducing points on $g$, we obtain the inducing points version: 
\begin{subequations}
  \label{app.eq16}
  \begin{align}
  \begin{split}
  \label{app.eq16a}
  &\mathbb{E}_q[\log p(\mathbf{y}^{r}\mid \{g_{i}^r\}_{i=1}^{I_r})]=\sum_{i=1}^{I_r}\sum_{n=1}^{N_i^r}-\log(\sigma_i\sqrt{2\pi})-\frac{1}{2\sigma_i^2}({y_{i,n}^{r^2}}-2y_{i,n}^r\bar{g}_{i,n}^r+{\tilde{g}_{i,n}^{r^2}})
  \end{split}\\
  \begin{split}
  \label{app.eq16b}
  &\text{KL}(q(\mathbf{g}_{\mathbf{x}_m})\Vert p(\mathbf{g}_{\mathbf{x}_m}))=\\
  &\frac{1}{2}\left(\log\lvert\mathbf{K}_{\mathbf{x}_m\mathbf{x}_m}\rvert-\log\lvert\mathbf{\Sigma}_{\mathbf{x}_m}\rvert-M\cdot I+\text{Tr}[\mathbf{K}_{\mathbf{x}_m\mathbf{x}_m}^{-1}\mathbf{\Sigma}_{\mathbf{x}_m}]+\mathbf{m}_{\mathbf{x}_m}^{\top}\mathbf{K}_{\mathbf{x}_m\mathbf{x}_m}^{-1}\mathbf{m}_{\mathbf{x}_m}\right),
  \end{split}
\end{align}
\end{subequations}
where we assume $p(\mathbf{g}_{\mathbf{x}_m})=\mathcal{N}(\mathbf{g}_{\mathbf{x}_m}\mid\mathbf{0},\mathbf{K}_{\mathbf{x}_m\mathbf{x}_m})$. Maximizing \cref{app.eq16a}, we obtain the optimal noise variance:
\begin{equation}
\sigma_i^{2*}=\left(\sum_{n=1}^{N_i^r}{y_{i,n}^{r^2}}-2y_{i,n}^r\bar{g}_{i,n}^r+{\tilde{g}_{i,n}^{r^2}}\right)/N_i^r.
\label{optimal_noise}
\end{equation}
Minimizing \cref{app.eq16b}, we obtain the optimal kernel hyperparameters $\{\bm{\theta}_q\}_{q=1}^Q$ and mixing weights $\{\mathbf{w}_q\}_{q=1}^Q$, which has no closed-form solution and we resort to the automatic differentiation technique.  
The pseudocode of mean-field approximation is provided in Algorithm~\ref{mf}. 
\begin{algorithm}
\caption{Mean-Field Approximation}\label{mf}
\begin{algorithmic}[1]
\State Initialize hyperparameters and variational parameters. 
\Repeat
    \State Update the optimal variational distribution of P\'{o}lya-Gamma variables for classification tasks in \cref{eq10a};
    \State Update the optimal variational distribution of P\'{o}lya-Gamma variables for Cox process tasks in \cref{eq10b};
    \State Update the optimal variational intensity of marked Poisson processes for Cox process tasks in \cref{eq10c};
    \State Update the optimal variational distribution of intensity upper-bounds for Cox process tasks in \cref{eq10d};
    \State Update the optimal variational distribution of latent functions for all tasks in \cref{eq10e};
    \State Update the hyperparameters $\{\bm{\theta}_q,\mathbf{w}_q\}_{q=1}^Q$ by minimizing \cref{app.eq16b};
    \State Update the hyperparameter $\sigma^2$ by \cref{optimal_noise}. 
\Until{convergence}\\
\Return $g_i^r(\mathbf{x})$ for regression task, $s(g_i^c(\mathbf{x}))$ for classification task and $\bar{\lambda}_i s(g_i^p(\mathbf{x}))$ for Cox process task. 
\end{algorithmic}
\end{algorithm}

Defining $S$ as the number of quadrature nodes on all point process tasks, the computational complexity of our mean-field approximation is dominated by the matrix inversion $O(M^3I^3)$ and product $O(M^2(N^r+N^c+N^p+S))$ where $N^{\cdot}$ is the number of samples in the corresponding tasks. 

\paragraph{Convergence and Minibatch}
The theoretical analysis in~\citet{hoffman2013stochastic} shows that performing the mean-field iteration for a conditionally conjugate model is equivalent to updating parameters by the natural gradient descent~\citep{amari1998natural} with a step size of one. Therefore, our proposed mean-field approximation has inherently a faster convergence than the standard gradient descent. 

The mean-field algorithm above uses all data. For further acceleration, we can resort to the stochastic variational inference~\citep{hoffman2013stochastic} by subsampling the tasks, and samples in regression and classification tasks.

\section{Experiments}
In this section, we analyze our model and inference on synthetic and real-world datasets to demonstrate the performance in terms of transfer capability, efficiency and convergence. For all experiments, we use the RBF kernel $k(\mathbf{x},\mathbf{x}')=\theta_0\exp{(-\frac{\theta_1}{2}\Vert\mathbf{x}-\mathbf{x}'\Vert^2)}$ as covariance functions, and the usage of other kernels is outside of the scope of this paper. The implementation code is publicly available at \url{https://github.com/zhoufeng6288/HGCox}. 

\paragraph{Baselines} To show the superiority of our approach, we compare our model HMGCP against the single-task Cox process model: variational LGCP~\citep{nguyen2014automated}, and the multi-task models: MLGCP~\citep{taylor2015bayesian} and MCPM~\citep{aglietti2019efficient}. 

\paragraph{Metrics} We provide the comparison result of our model with baselines in terms of \emph{estimation error (EE)}, \emph{test log-likelihood (TLL)}, \emph{running time (RT)} and \emph{convergence rate (CR)}. EE is the root mean square error (RMSE) between the estimated parameter and the ground truth. 
It is worth noting that EE is only applicable to synthetic data because the ground truth is required. 
TLL is the log-likelihood on test data using the posterior mean of parameters estimated from training data. RT is the running time of the inference algorithm. CR is the convergence rate of training log-likelihood w.r.t. the number of iterations.


\subsection{Synthetic Data: Complete}
\label{sec5.1}

To illustrate the performance of transfer capability, efficiency and convergence of our approach, we simulate three heterogeneous correlated tasks (one regression, one binary classification and one Cox process) by sampling three latent functions from a MOGP prior and using them to simulate the observed samples in regression, classification and Cox process tasks. 
We simulate three sets of synthetic data using three different sets of hyperparameters where latent functions vary from gently to drastically; each synthetic dataset contains both training and test data. We use two basis functions. The hyperparameters are $\sigma^2=0.1$, $\bm{\theta}_1=[1,0.001]$, $\bm{\theta}_2=[1,0.001]$, $\mathbf{w}_1=[0.9,0.5,0.1]$ and $\mathbf{w}_2=[0.1,0.5,0.9]$ for the first dataset; $\sigma^2=0.1$, $\bm{\theta}_1=[1,0.02]$, $\bm{\theta}_2=[2,0.001]$, $\mathbf{w}_1=[0.9,0.5,0.1]$ and $\mathbf{w}_2=[0.1,0.5,0.9]$ for the second dataset; $\sigma^2=0.1$, $\bm{\theta}_1=[1,0.1]$, $\bm{\theta}_2=[2,0.1]$, $\mathbf{w}_1=[0.9,0.5,0.1]$ and $\mathbf{w}_2=[0.1,0.5,0.9]$ for the third dataset. 

For each dataset, we draw two basis functions $\{f_q\}_{q=1}^2$ on the domain $[0,100]$ from two independent zero-mean GP priors with the corresponding kernel hyperparameters. The task-specific latent functions are $\{g_i=\sum_{q=1}^2w_{i,q}f_q\}_{i=1}^3$. $g_1$ is used as the mean of a Gaussian distribution $\mathcal{N}(g_1(\mathbf{x}),\sigma^2)$ to draw samples for the regression task. $g_2$ is passed through a sigmoid function and then used as the parameter of a Bernoulli distribution to draw samples for the binary classification task. $g_3$ is passed through a sigmoid function and then scaled by $\bar{\lambda}$ to serve as the intensity for simulating a Cox process. For regression and classification tasks, we assume the samples are uniformly distributed on the domain. 

\begin{figure}[ht]
\begin{center}
\begin{minipage}{0.65\columnwidth}
\includegraphics[width=\columnwidth]{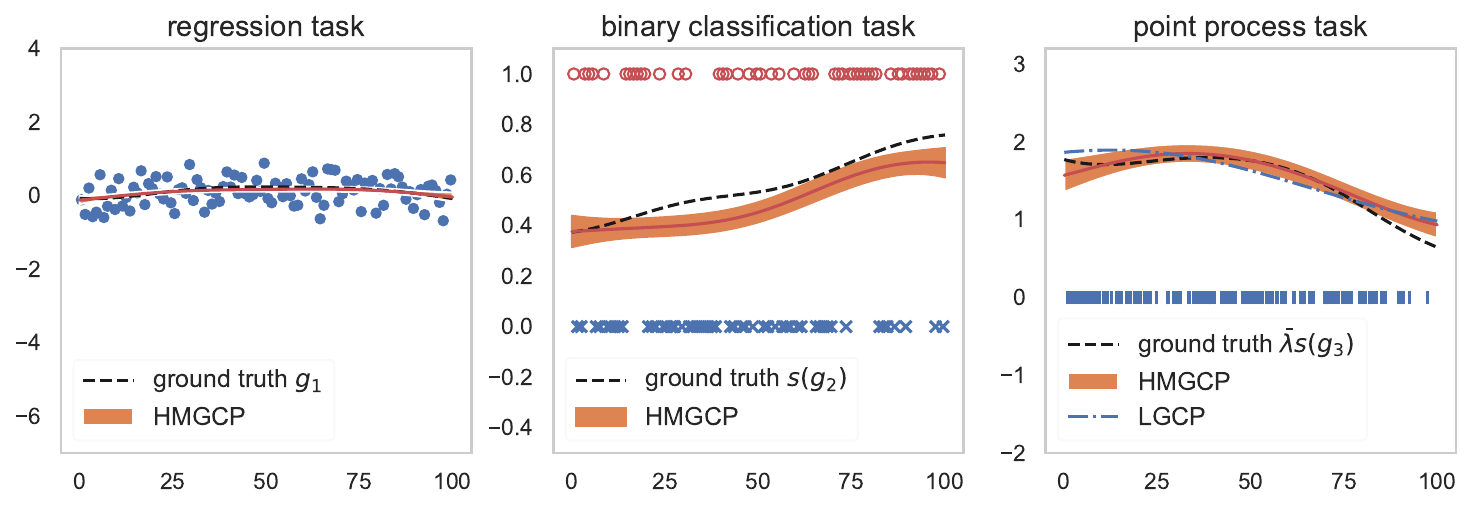}
\subcaption{First Dataset (latent functions vary gently)}\label{app.fig1a}
\end{minipage}
\begin{minipage}{0.65\columnwidth}
\includegraphics[width=\columnwidth]{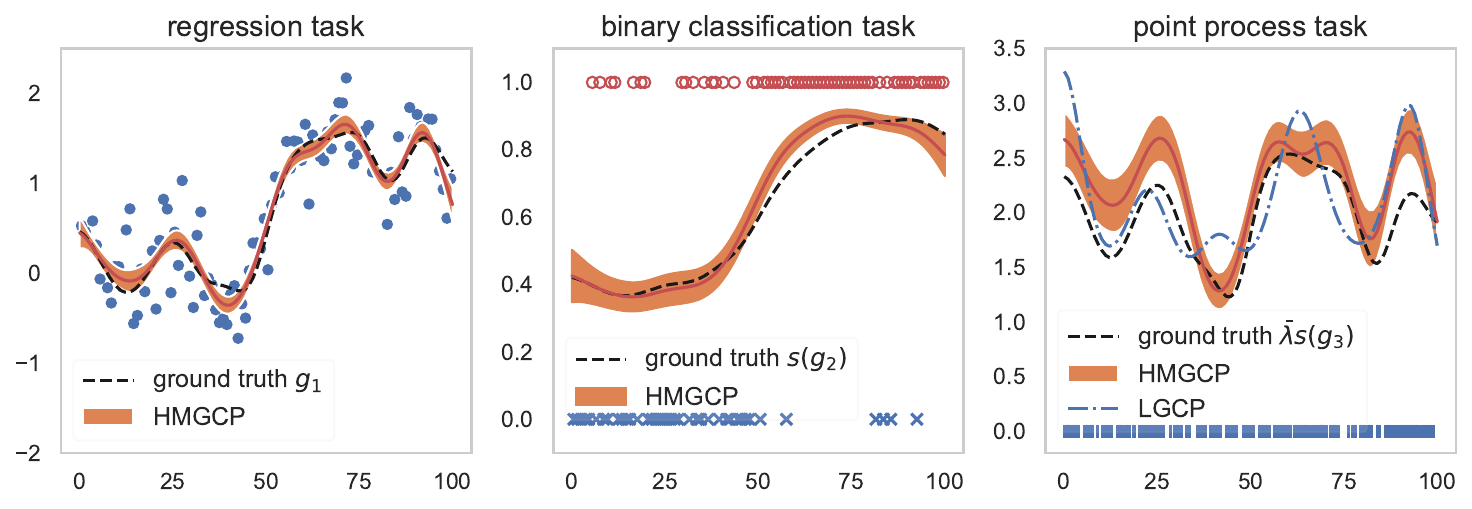}
\subcaption{Second Dataset (latent functions vary moderately)}\label{app.fig1b}
\end{minipage}
\begin{minipage}{0.65\columnwidth}
\includegraphics[width=\columnwidth]{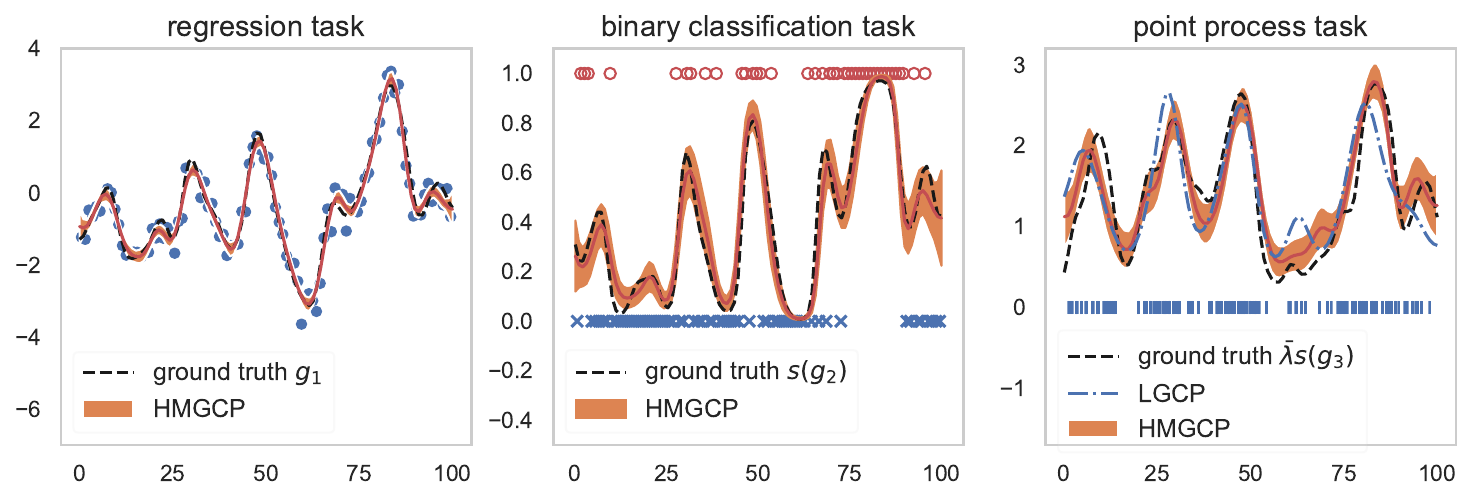}
\subcaption{Third Dataset (latent functions vary drastically)}\label{app.fig1c}
\end{minipage}
\caption{HMGCP recovers the latent functions $g_1$, $s(g_2)$ and $\bar{\lambda}s(g_3)$ in three datasets whose posterior is constructed by 100 samples of $g$ and $\bar{\lambda}$ from the corresponding variational posterior. The shading area indicates one standard deviation. Blue dots are samples in regression task; red circles and blue crosses are positive and negative samples in classification task; blue bars are samples in Cox process task. For LGCP, we show the posterior mean intensity for the Cox process task.}
\label{fig1}
\end{center}
\end{figure}

Our goal is to recover the intensity upper-bound $\bar{\lambda}$ and latent functions $\{g_i\}_{i=1}^3$. We use $30$ inducing points that are uniformly distributed on the domain and $100$ Gaussian quadrature nodes for the intractable integral. For initialization, the initial hyperparameters $\sigma^2$, $\{\bm{\theta}_q,\mathbf{w}_q\}_{q=1}^2$ are set to the ground-truth hyperparameters and the variational parameters are initialized randomly. In the training process, the variational parameters and hyperparameters are updated concurrently.
Specifically, the variational parameters are updated by the mean-filed iteration, the kernel hyperparameters $\{\bm{\theta}_q,\mathbf{w}_q\}_{q=1}^2$ are updated by minimizing \cref{app.eq16b} using the `SLSQP' method, and the noise variance $\sigma^2$ is updated by \cref{optimal_noise}. 
\Cref{fig1} represents the estimated result for three datasets where we can see HMGCP is able to recover the ground truth. For convergence, HMGCP only takes 2-3 steps to converge in terms of training log-likelihood, which is much faster than the first-order gradient-based LGCP requiring more than 500 steps. More importantly, HMGCP has the better EE and TLL (\cref{tab1}) than the single-task LGCP that is trained independently and not able to transfer information to help recover the intensity of Cox process. For a fair comparison of efficiency, we run both HMGCP and LGCP on a single Cox process task with 400 iterations, and our inference is at least twice as fast as LGCP (\cref{tab1}) demonstrating its outstanding efficiency. 

\begin{table}[t]
\caption{The performance of EE, TLL and RT for HMGCP and LGCP on three synthetic datasets. EE is the RMSE between posterior mean and ground truth. Time in seconds.}
\label{tab1}
\begin{center}
\scalebox{0.8}{
\begin{sc}
\begin{tabular}{lcccccccc}
\toprule
& Model & EE(reg) & EE(cla) & EE(Cox) & TLL(reg) & TLL(cla) & TLL(Cox) & RT\\
\midrule
\multirow{2}{*}{1} & HMGCP & 0.046 & 0.074 & \textbf{0.114} & -33.17 & -63.57 & \textbf{-89.05} & \textbf{0.73}\\
& LGCP & $\times$ & $\times$ & 0.147 & $\times$ & $\times$ & -90.23 & 2.70\\
\midrule
\multirow{2}{*}{2} & HMGCP & 0.098 & 0.048 & \textbf{0.319} & -28.54 & -55.23 & \textbf{-63.54} & \textbf{1.09}\\
& LGCP & $\times$ & $\times$ & 0.385 & $\times$ & $\times$ & -65.19 & 2.73\\
\midrule
\multirow{2}{*}{3} & HMGCP & 0.167 & 0.067 & \textbf{0.272} & -42.43 & -56.14 & \textbf{-72.75} & \textbf{0.69}\\
& LGCP & $\times$ & $\times$ & 0.433 & $\times$ & $\times$ & -79.17 & 2.71\\
\bottomrule
\end{tabular}
\end{sc}}
\end{center}
\end{table}

\subsection{Synthetic Data: Missing}
\label{sec5.2}
As far as we know, all current multi-task Cox process models exclusively focus on \emph{homogeneous} scenarios. This does not apply to the more general \emph{heterogeneous} multi-task setup where we need to transfer knowledge between multiple heterogeneous correlated tasks. In this section, we compare HMGCP against homogeneous multi-task baselines: MLGCP and MCPM. We construct four heterogeneous correlated tasks (one regression, one binary classification and two Cox processes) using the same method as in \cref{sec5.1}. We simulate one set of synthetic data that contains both training and test data. We use two basis functions. The hyperparameters are $\sigma^2=0.1$, $\bm{\theta}_1=[1,0.02]$, $\bm{\theta}_2=[2,0.001]$, $\mathbf{w}_1=[0.9,0.1,0.3,1.0]$ and  $\mathbf{w}_2=[0.1,0.9,0.5,1.0]$. To further illustrate the heterogeneous transfer capability of our approach, in addition to the complete data, we follow the experimental setup of~\citet{aglietti2019efficient}: we create some \emph{missing gaps} by evenly partitioning the domain into several regions and randomly masking four non-overlapping regions on four tasks (one for each task). To demonstrate the transfer capability on problems with different levels of difficulty, we experiment with two missing-gap widths: $5$ and $10$, where a wider missing gap means a more difficult transfer problem. For each missing-gap width, we experiment with ten random configurations of missing gaps.

\begin{figure}[htbp]
\begin{center}
\begin{minipage}{\textwidth}
\centering
    \begin{minipage}[b]{0.42\columnwidth}
    \includegraphics[width=\columnwidth]{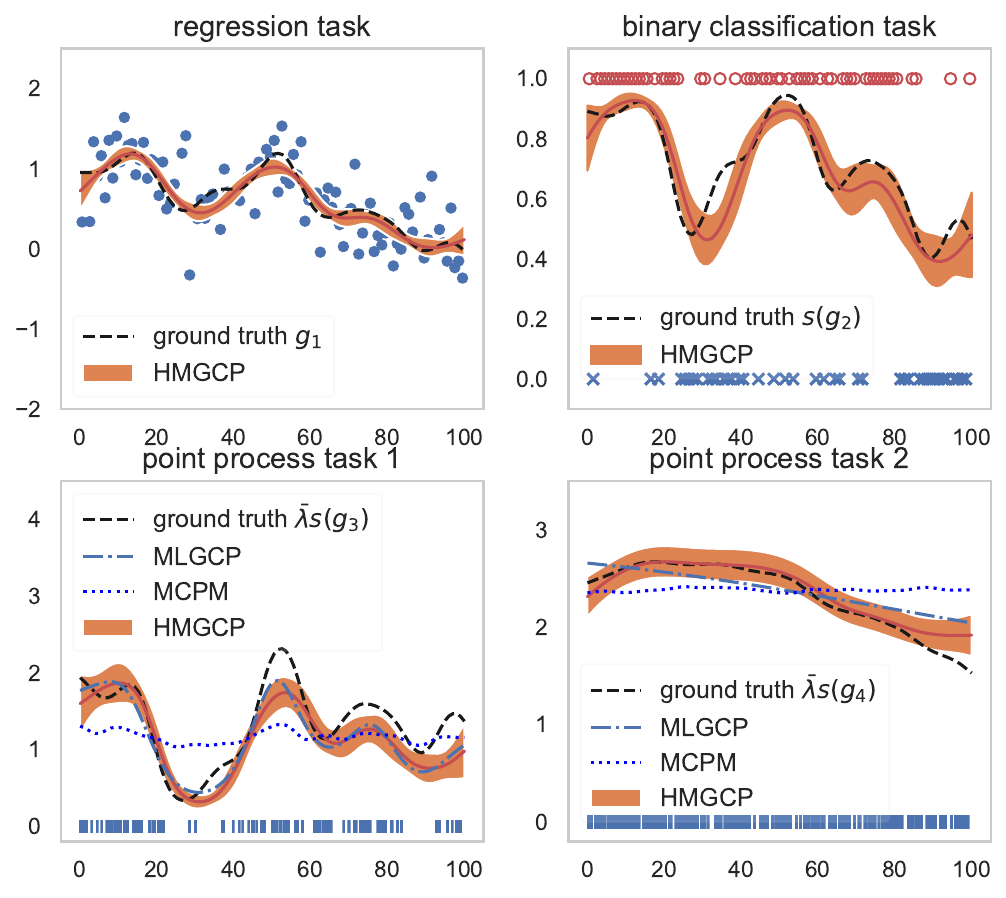}
    \end{minipage}
\subcaption{Missing-Gap Width: 0 (Complete Data)}
\end{minipage}
\begin{minipage}{\textwidth}
\centering
    \begin{minipage}[b]{0.42\columnwidth}
    \includegraphics[width=\columnwidth]{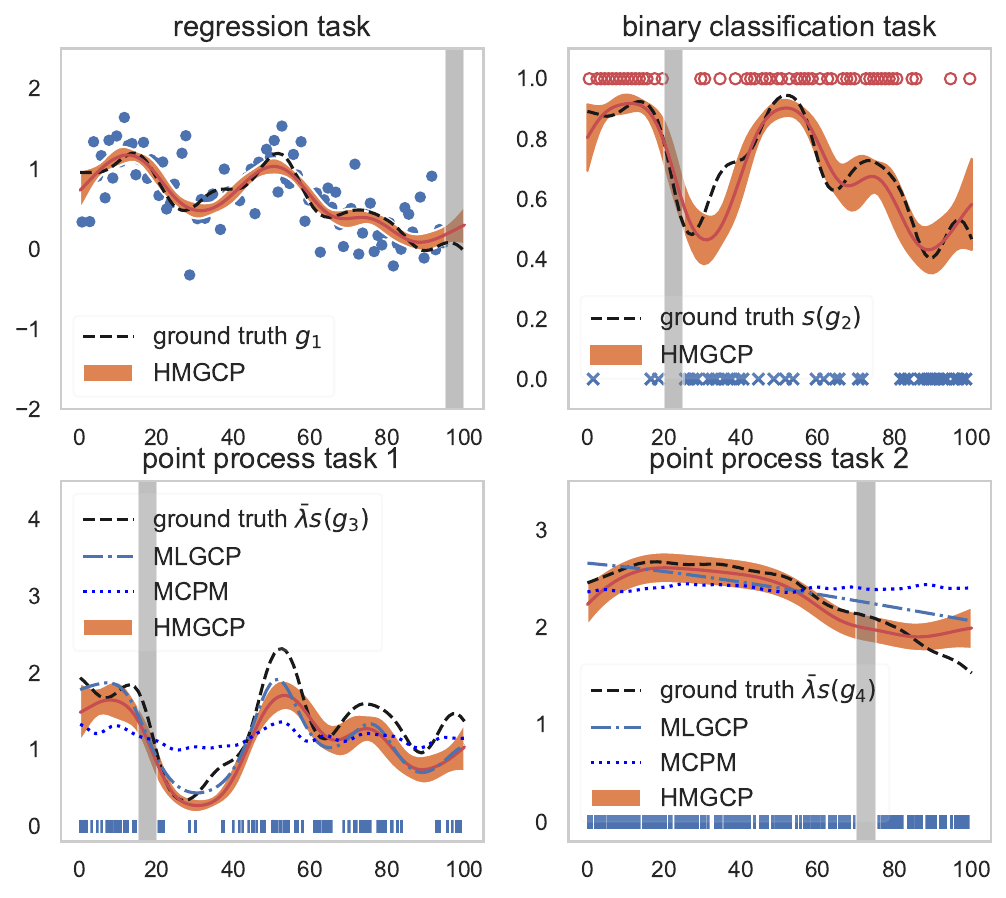}
    \end{minipage}
    \begin{minipage}[b]{0.42\columnwidth}
    \includegraphics[width=\columnwidth]{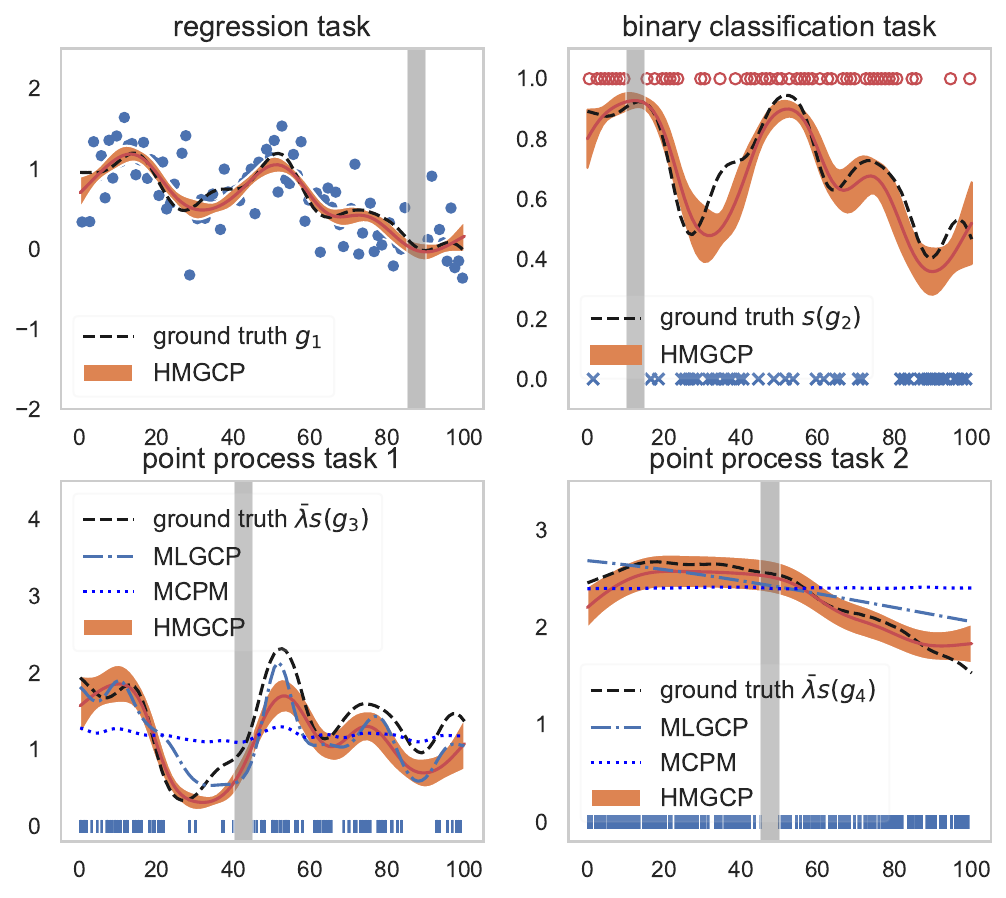}
    \end{minipage}
\subcaption{Missing-Gap Width: 5}
\end{minipage}
\begin{minipage}{\textwidth}
\centering
    \begin{minipage}[b]{0.42\columnwidth}
    \includegraphics[width=\columnwidth]{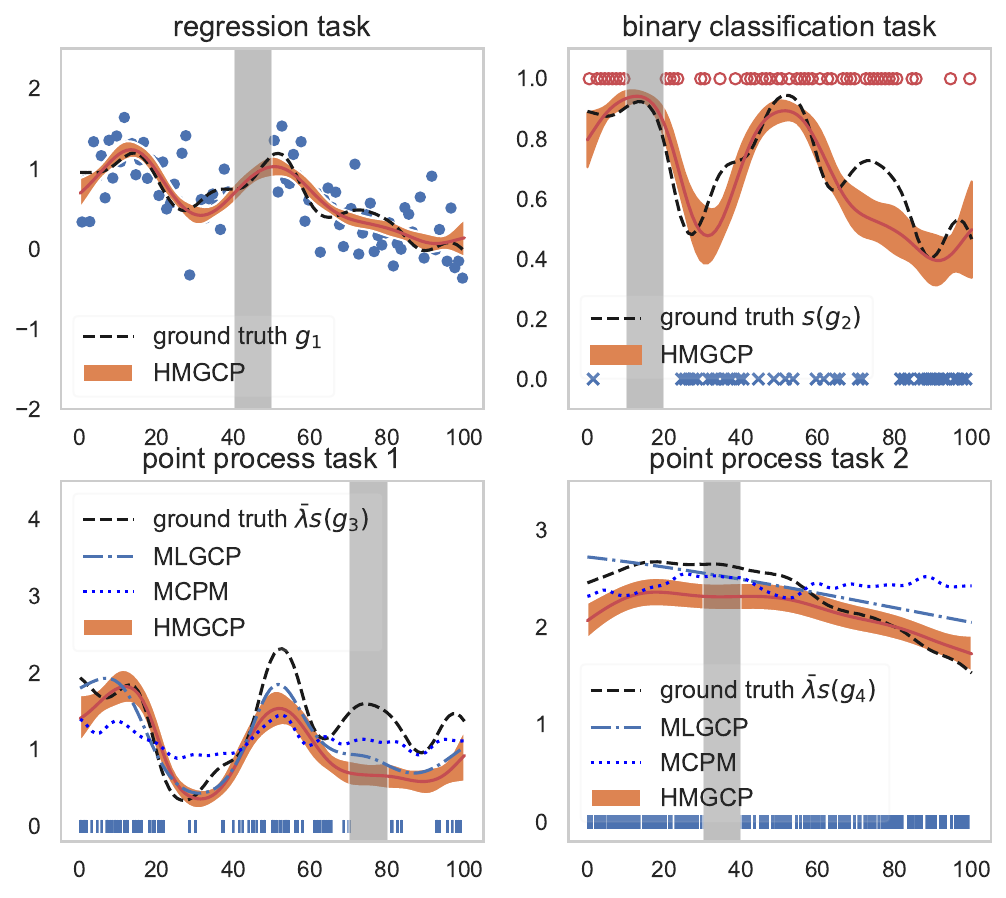}
    \end{minipage}
    \begin{minipage}[b]{0.42\columnwidth}
    \includegraphics[width=\columnwidth]{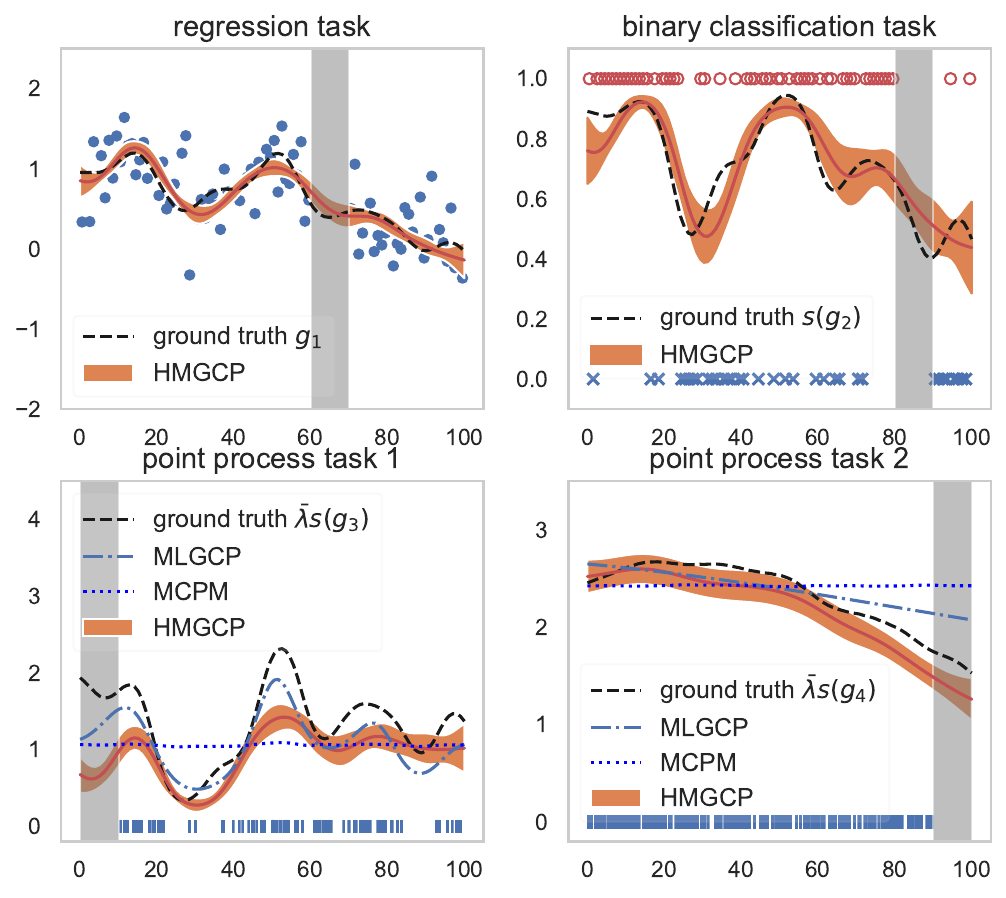}
    \end{minipage}
\subcaption{Missing-Gap Width: 10}
\end{minipage}
\caption{The estimated posterior of latent functions $g_1$, $s(g_2)$, $\bar{\lambda}_3s(g_3)$ and $\bar{\lambda}_4s(g_4)$ from HMGCP with missing-gap width being (a) $0$, (b) $5$ and (c) $10$. For missing-gap widths $5$ and $10$, we show two configurations of missing gaps across tasks. The grey areas indicate the masked missing gaps. For MLGCP and MCPM, we show the posterior mean intensities for two Cox process tasks. The posterior variance in the missing gap does not increase significantly meaning HMGCP successfully transfers heterogeneous knowledge.}
\label{fig2}
\end{center}
\end{figure}

We use $10$ inducing points which are uniformly distributed on the domain. All the other experimental settings are the same as in \cref{sec5.1}. HMGCP successfully transfers knowledge between heterogeneous tasks by exploiting commonalities between them to recover the missing-gap latent functions for all tasks (\cref{fig2}), whereas MLGCP and MCPM exhibit the inferior generalization capability since they can only share information between Cox processes. \Cref{fig2} shows the estimated latent functions for several configurations with 3 different missing-gap widths across tasks. Generally, the transfer of knowledge in regression and classification tasks is easier than that in Cox process tasks. This is because the likelihood of regression and classification only considers observed points, the function in the missing gap is entirely determined by the smoothness induced by prior. However, in addition to observed points, the Cox process likelihood also considers the domain where no points appear, so the function in the missing gap is determined by both prior and likelihood (zero-valued intensity). This makes the estimated intensity in the missing gap generally lower than the ground truth. For each missing-gap width, we report the statistics of EE and TLL for HMGCP, MLGCP and MCPM over ten random configurations of missing gaps in~\cref{tab2} where HMGCP outperforms alternatives in all experiments. The reason is HMGCP extracts useful information from regression, classification and other Cox processes to improve the estimation of intensity for the current Cox process, while MLGCP and MCPM cannot incorporate the information existing in heterogeneous tasks. As in~\cref{sec5.1}, we run HMGCP, MLGCP and MCPM only on the complete Cox process data for a fair comparison of efficiency: HMGCP consumes $3.68$ seconds, while MLGCP and MCPM consume $12.15$ and $21.36$ seconds, respectively (2000 iterations). 

\begin{table}[t]
\caption{The performance of EE and TLL for HMGCP, MLGCP and MCPM over ten random configurations of missing gaps with three different missing-gap widths ($0$ means complete data). The mean and standard deviation (in brackets) are provided. EE(Cox)/TLL(Cox) is the sum of EEs/TLLs of two Cox processes.}
\label{tab2}
\centering
\scalebox{0.7}{
\begin{sc}
\begin{tabular}{cccccccc}
\toprule
Gap Width & Model & EE(reg) & EE(cla) & EE(Cox) & TLL(reg) & TLL(cla) & TLL(Cox)\\
\midrule
\multirow{3}{*}{0} & HMGCP & 0.093 & 0.066 & \textbf{0.390} & -50.61 & -56.67 & \textbf{-120.55}\\
& MLGCP & $\times$ & $\times$ & 0.535 & $\times$ & $\times$ & -136.28\\
& MCPM & $\times$ & $\times$ & 0.676 & $\times$ & $\times$ & -126.73\\
\midrule
\multirow{3}{*}{5} & HMGCP & 0.095(0.006) & 0.066(0.005) & \textbf{0.461}(0.056) & -50.76(0.92) & -56.74(0.51) & \textbf{-122.94}(2.27)\\
& MLGCP & $\times$ & $\times$ & 0.601(0.051) & $\times$ & $\times$ & -126.24(3.39) \\
& MCPM & $\times$ & $\times$ & 0.725(0.035) & $\times$ & $\times$ & -129.82(2.53)\\
\midrule
\multirow{3}{*}{10} & HMGCP & 0.111(0.006) & 0.072(0.008) & \textbf{0.664}(0.071) & -52.14(1.94) & -56.82(0.69) & \textbf{-128.49}(5.74)\\
& MLGCP & $\times$ & $\times$ & 0.791(0.070) & $\times$ & $\times$ & -128.59(5.33)\\
& MCPM & $\times$ & $\times$ & 0.765(0.024) & $\times$ & $\times$ & -131.59(1.99)\\
\bottomrule
\end{tabular}
\end{sc}} 
\end{table}

\subsection{Real Data}
\label{sec5.3}
\begin{figure}[t]
\centering
\includegraphics[width=0.7\columnwidth]{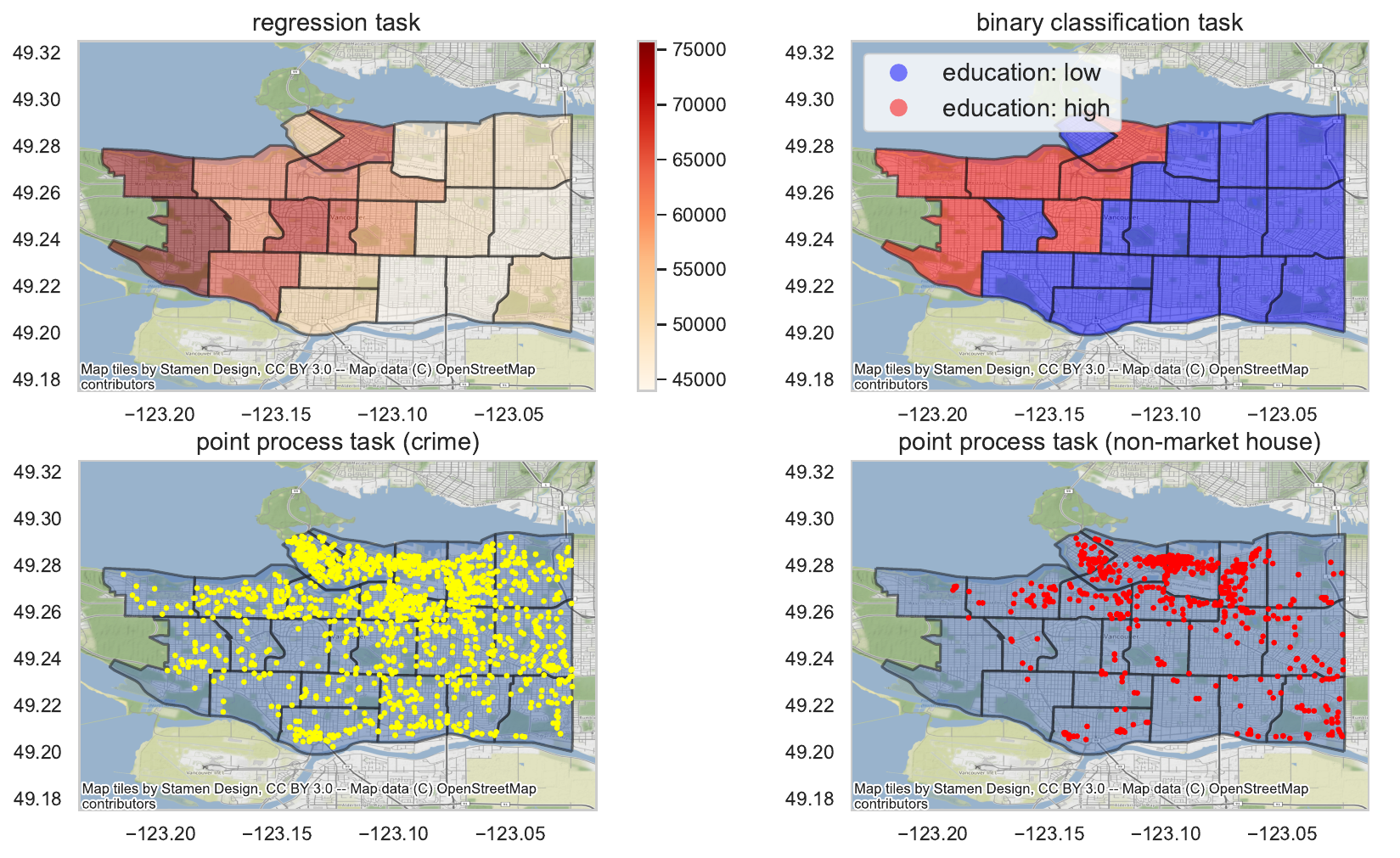}
\caption{The median employment income (top left), education degree (top right), theft of vehicle (bottom left) and non-market house (bottom right) in $22$ neighbourhoods of Vancouver.}
\label{fig3}
\end{figure}

In this section, we demonstrate the superiority of HMGCP in terms of heterogeneous knowledge transfer, efficiency and convergence on a real-world 2D urban data of Vancouver. The dataset\footnote{The income, education and non-market housing data is from the Vancouver Open Data Catalog (https://opendata.vancouver.ca/pages/home/). The crime data is from Kaggle (https://www.kaggle.com/datasets/wosaku/crime-in-vancouver).} contains four parts of data (\cref{fig3}): 
\textbf{(1)} \emph{Employment income in Vancouver}: the median employment income for full-year full-time workers in 2015 in the neighbourhoods of Vancouver; 
\textbf{(2)} \emph{Education in Vancouver}: the number of population holding university certificate, diploma or degree at bachelor level or above in the neighbourhoods of Vancouver; 
\textbf{(3)} \emph{Crime in Vancouver}: the recording of miscellaneous crimes (type, neighbourhood, latitude, longitude) in 2015 in Vancouver; 
\textbf{(4)} \emph{Non-market housing in Vancouver}: the information of non-market housing projects (name, address, neighbourhood, latitude, longitude) that is for low and moderate income singles and families.

For the first dataset, we formulate it as a regression task, and use the centroid of each neighbourhood as the input, the median income as the output; for the second dataset, we formulate it as a binary classification task according to the degree of education: we divide the $22$ neighbourhoods into `$+1$' if there are more people holding university certificate, diploma or degree at bachelor level or above, and `$-1$' if not; for the third and fourth datasets, we extract the locations of `Theft of Vehicle' records in 2015 and non-market housing projects respectively, and formulate them as two Cox process tasks. On the basis of common sense, the income level, education degree, crime rate and non-market housing are closely correlated. Therefore, their integrative analysis offers more advantages compared to learning multiple tasks independently, which is susceptible to overfitting.

\begin{figure}[htbp]
\begin{center}
\begin{minipage}{\textwidth}
\centering
    \begin{minipage}[b]{0.7\columnwidth}
    \includegraphics[width=\columnwidth]{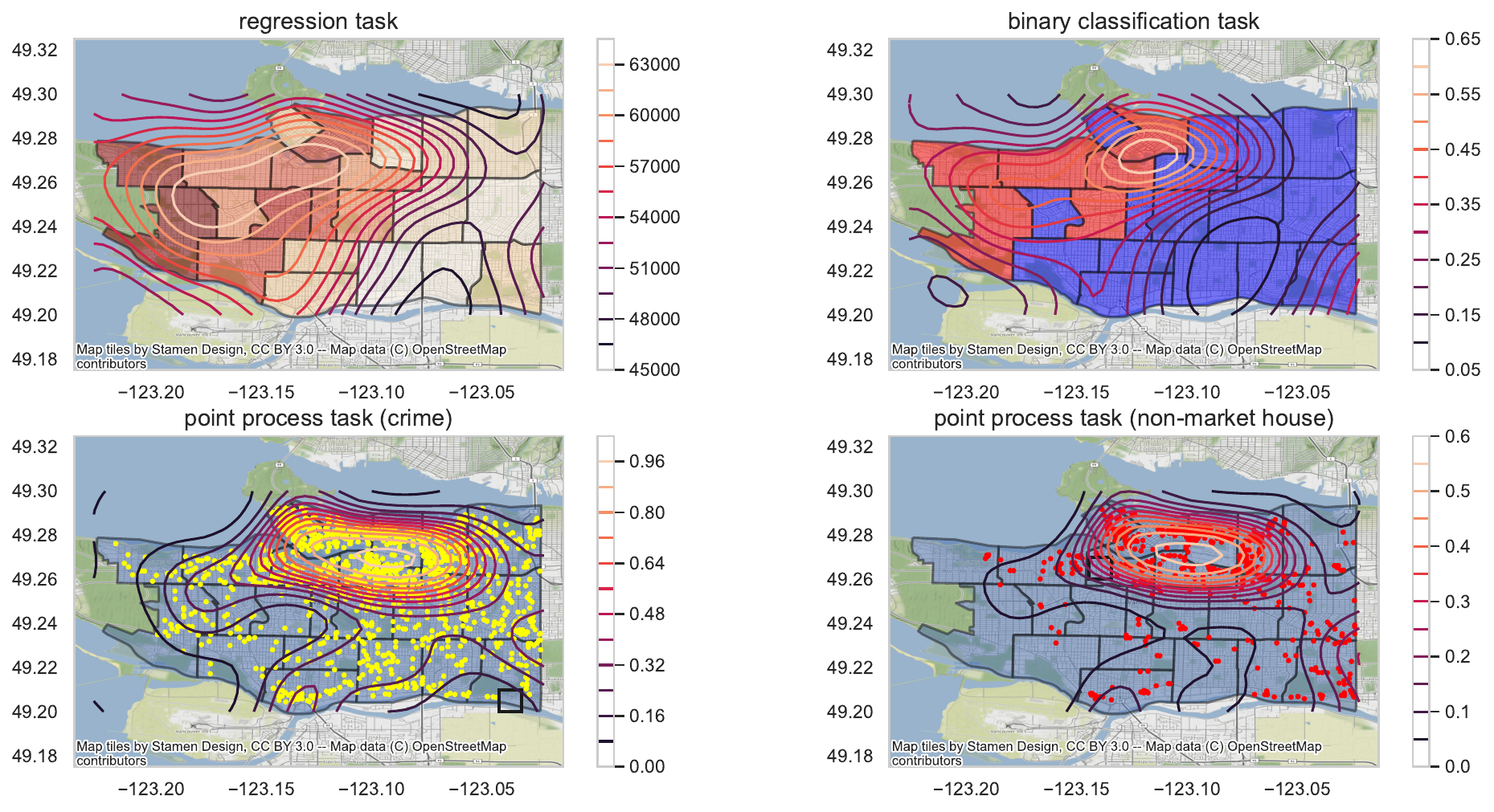}
    \end{minipage}
\subcaption{Size of mask: $5\times5$}
\end{minipage}
\begin{minipage}{\textwidth}
\centering
    \begin{minipage}[b]{0.7\columnwidth}
    \includegraphics[width=\columnwidth]{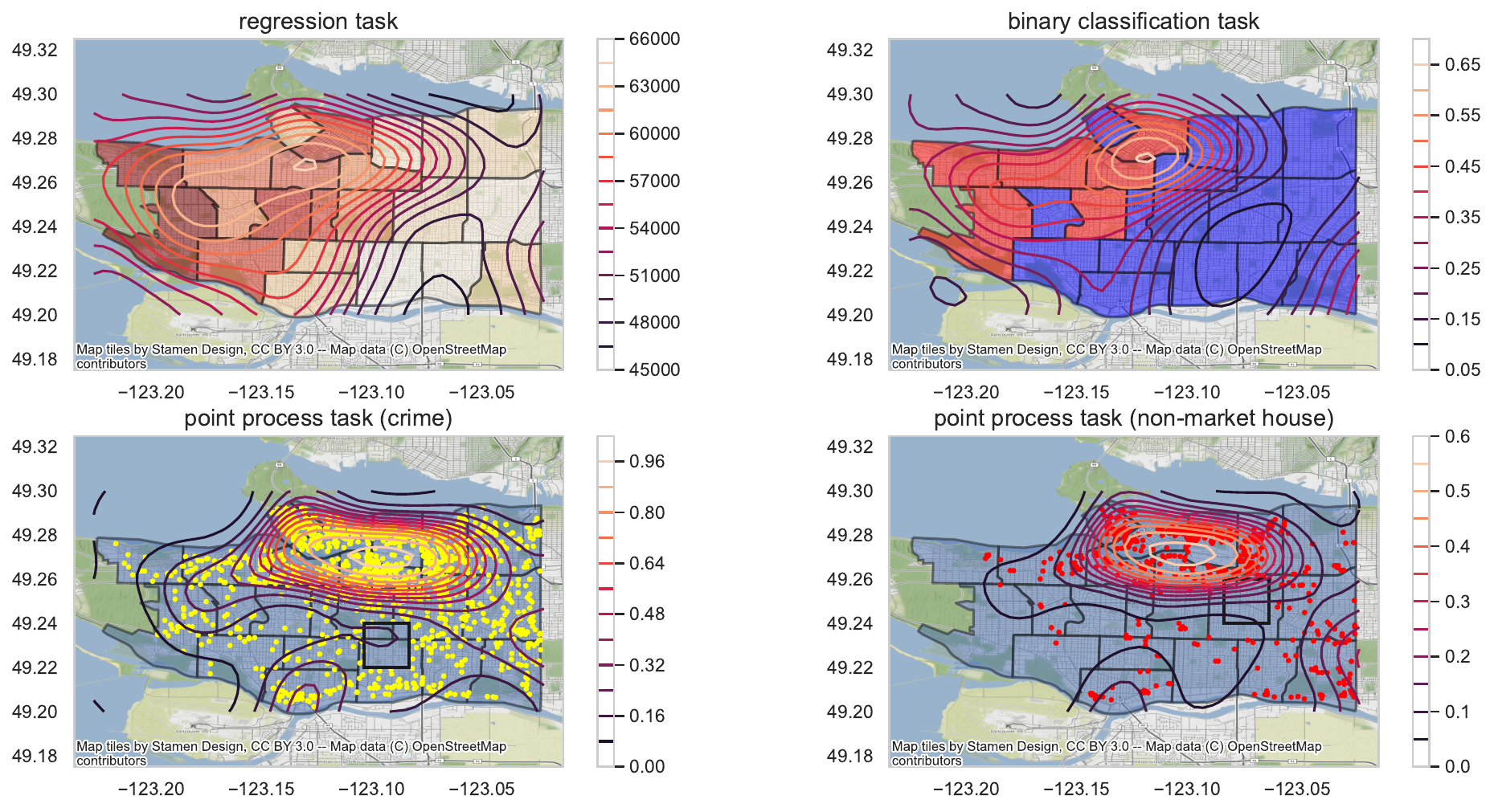}
    \end{minipage}
\subcaption{Size of mask: $10\times10$}
\end{minipage}
\begin{minipage}{\textwidth}
\centering
    \begin{minipage}[b]{0.7\columnwidth}
    \includegraphics[width=\columnwidth]{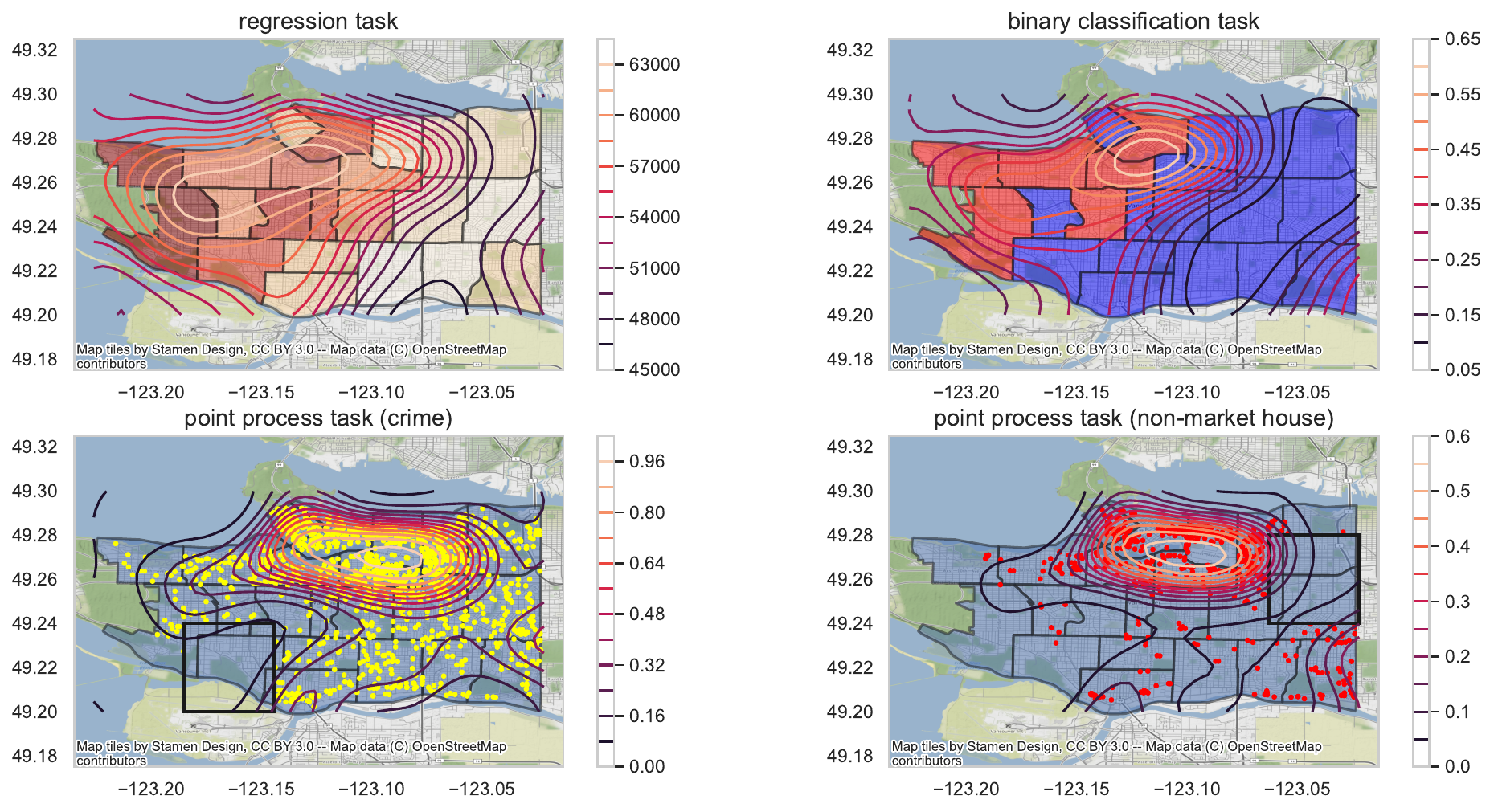}
    \end{minipage}
\subcaption{Size of mask: $20\times20$}
\end{minipage}
\caption{The estimated posterior mean latent functions $g_1$, $s(g_2)$, $\bar{\lambda}_3s(g_3)$ and $\bar{\lambda}_4s(g_4)$ from HMGCP with the mask size being (a) $5\times5$, (b) $10\times10$ and (c) $20\times20$ on each Cox process task. We show one configuration of masked regions across Cox process tasks. The black boxes indicate the masked regions.}
\label{app.fig3}
\end{center}
\end{figure}

\begin{figure}[htbp]
\begin{center}
\begin{minipage}{\textwidth}
\centering
    \begin{minipage}[b]{0.7\columnwidth}
    \includegraphics[width=\columnwidth]{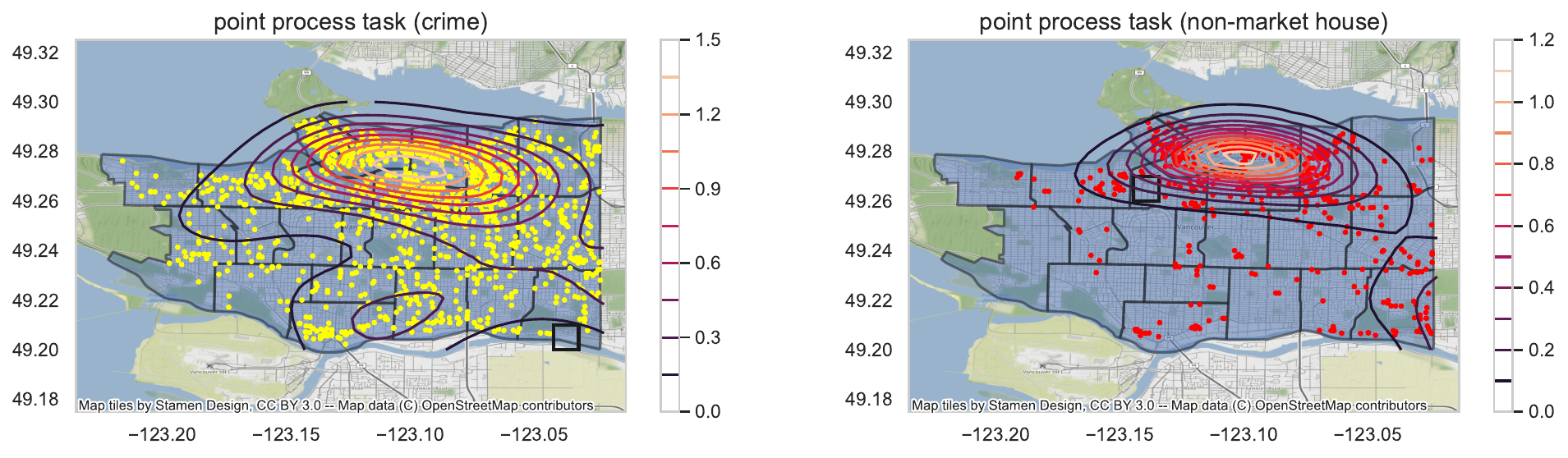}
    \end{minipage}
    \begin{minipage}[b]{0.7\columnwidth}
    \includegraphics[width=\columnwidth]{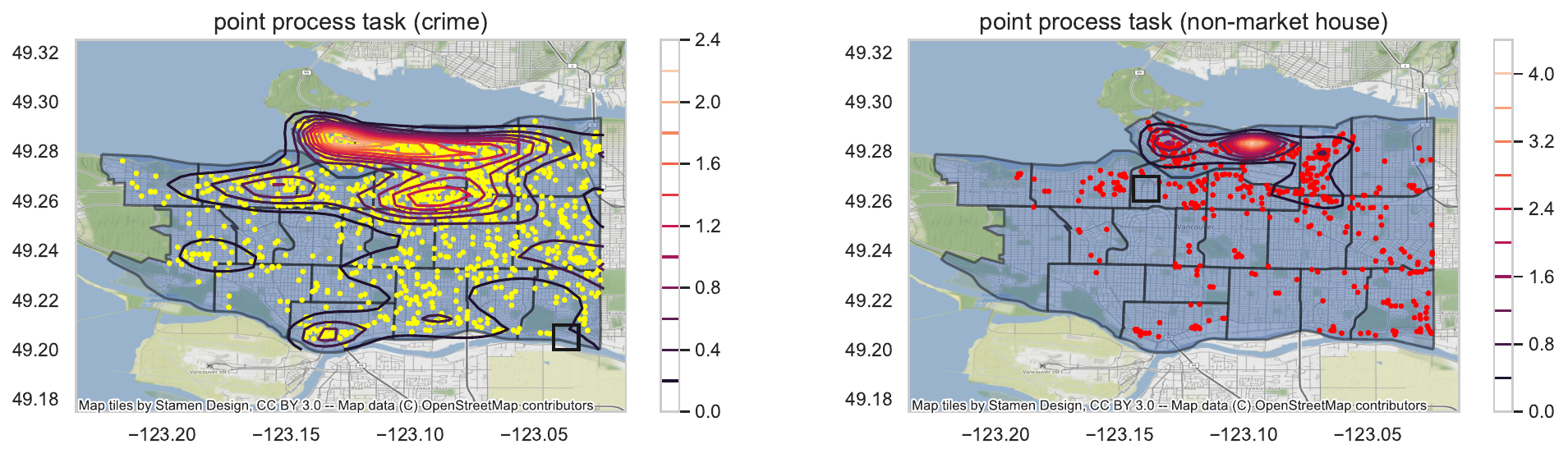}
    \end{minipage}
\subcaption{Size of mask: $5\times5$ (top: MLGCP, bottom: MCPM)}
\end{minipage}
\begin{minipage}{\textwidth}
\centering
    \begin{minipage}[b]{0.7\columnwidth}
    \includegraphics[width=\columnwidth]{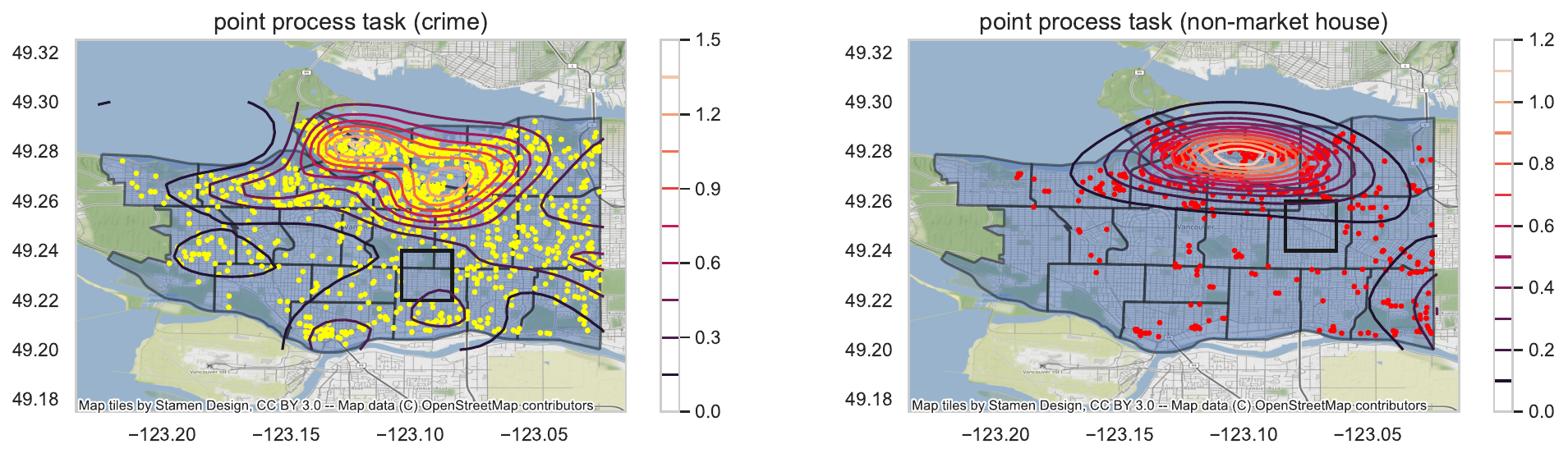}
    \end{minipage}
    \begin{minipage}[b]{0.7\columnwidth}
    \includegraphics[width=\columnwidth]{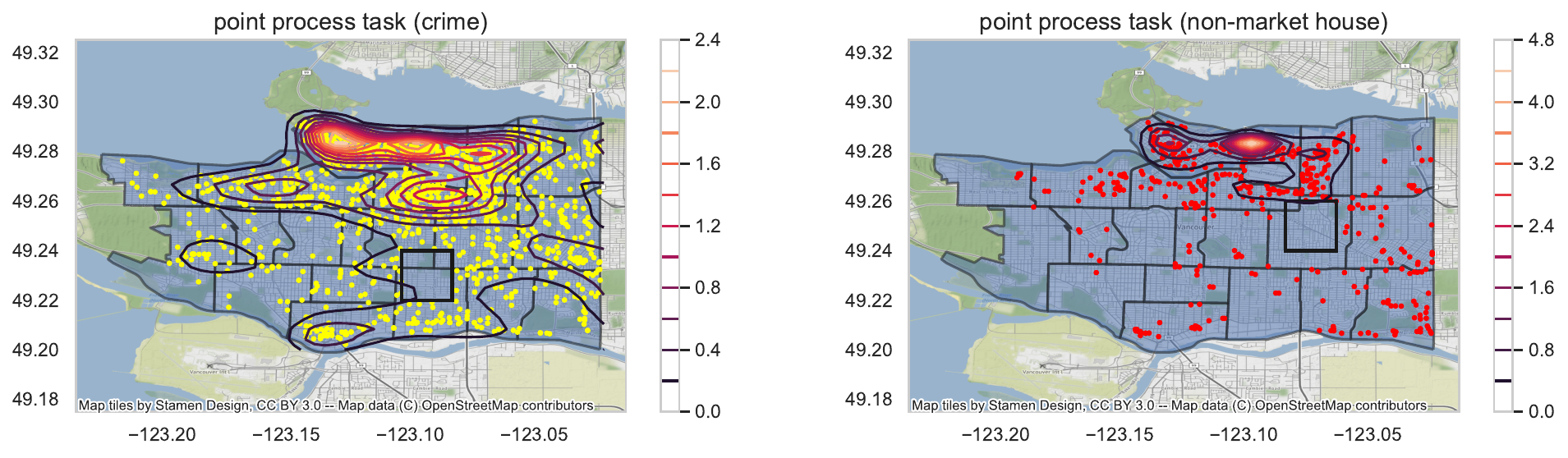}
    \end{minipage}
\subcaption{Size of mask: $10\times10$ (top: MLGCP, bottom: MCPM)}
\end{minipage}
\begin{minipage}{\textwidth}
\centering
    \begin{minipage}[b]{0.7\columnwidth}
    \includegraphics[width=\columnwidth]{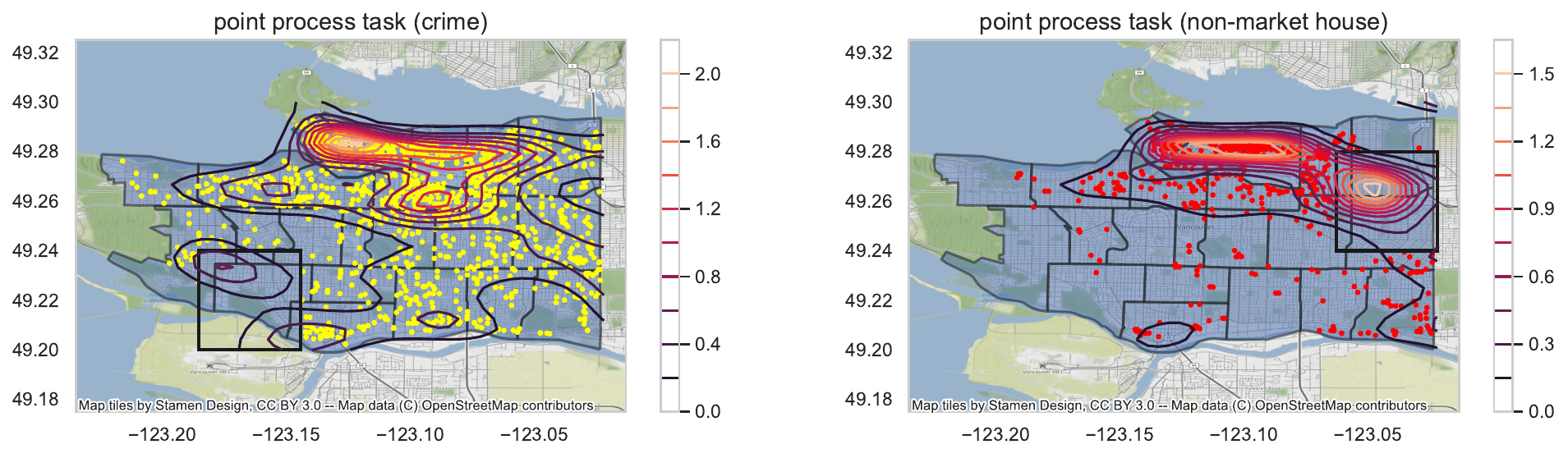}
    \end{minipage}
    \begin{minipage}[b]{0.7\columnwidth}
    \includegraphics[width=\columnwidth]{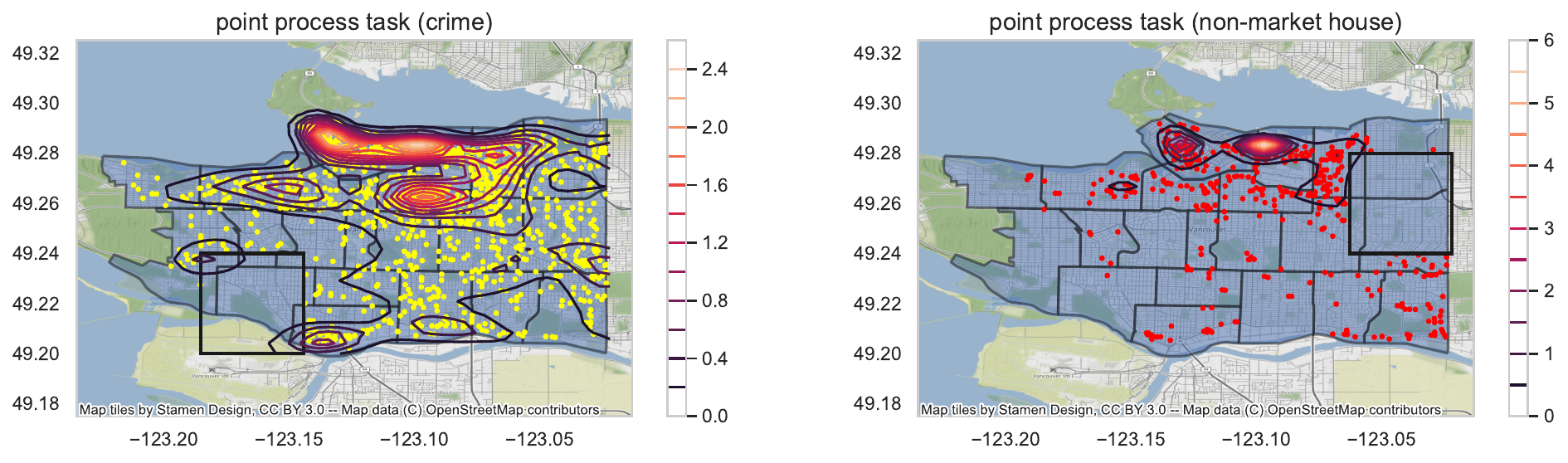}
    \end{minipage}
\subcaption{Size of mask: $20\times20$ (top: MLGCP, bottom: MCPM)}
\end{minipage}
\caption{The estimated posterior mean intensity functions for two Cox process tasks from MLGCP and MCPM with the mask size being (a) $5\times5$, (b) $10\times10$ and (c) $20\times20$ on each Cox process task. We show one configuration of masked regions across Cox process tasks. The black boxes indicate the masked regions.}
\label{app.fig4}
\end{center}
\end{figure}

To show the heterogeneous transfer capability of our approach, we compare HMGCP against MLGCP and MCPM. Due to lack of ground-truth latent functions, we cannot compare them in terms of EE but only TLL. We scale the area of Vancouver between longitude $[-123.226,-123.022]$ and latitude $[49.20, 49.30]$ to the domain $[0,100]\times[0,50]$. 
We choose three basis functions by trial and error: we gradually increase the number of basis functions and find that using three basis functions can achieve excellent performance. Using more basis functions only has a slight impact on the performance on the test data, but leads to longer training time. 
The initial hyperparameters are set to $\sigma^2=0.1$, $\bm{\theta}_1=[1,0.01]$, $\bm{\theta}_2=[1,0.005]$, $\bm{\theta}_3=[1,0.001]$, $\mathbf{w}_1=[0.5,0.5,0.1,0.1]$, $\mathbf{w}_2=[0.1,0.5,0.2,0.5]$ and $\mathbf{w}_3=[0.5,0.1,0.5,0.2]$, and the variational parameters are initialized randomly. 
In the training process, the variational parameters and hyperparameters are updated concurrently.
Specifically, the variational parameters are updated by the mean-filed iteration, the kernel hyperparameters $\{\bm{\theta}_q,\mathbf{w}_q\}_{q=1}^3$ are updated by minimizing \cref{app.eq16b} using the `SLSQP' method, and the noise variance $\sigma^2$ is updated by \cref{optimal_noise}. 
To assess the transfer capability with different levels of difficulty, we follow the experimental setup in~\cref{sec5.2}: we randomly mask two non-overlapping regions on \emph{Crime in Vancouver} and \emph{Non-market housing in Vancouver}, one for each task, with three different mask sizes: $5\times5$, $10\times10$ and $20\times20$. A larger mask indicates a more difficult transfer problem. For each mask size, we experiment with ten random configurations of masks.

\begin{figure}[htbp]
\begin{center}
\begin{minipage}{0.31\textwidth}
\centering
\includegraphics[width=\columnwidth]{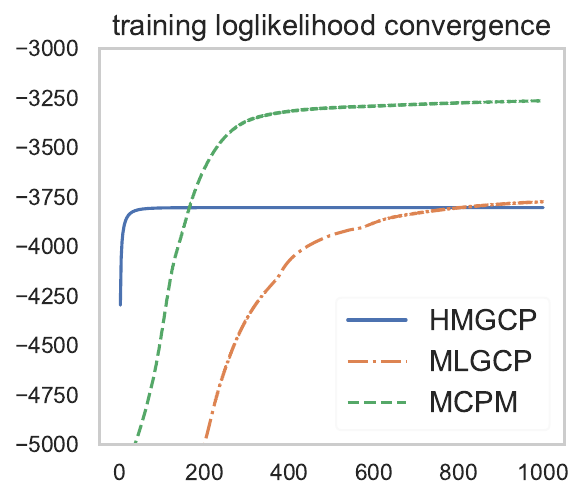}
\subcaption{Size of mask: $5\times5$}
\end{minipage}
\begin{minipage}{0.31\textwidth}
\centering
\includegraphics[width=\columnwidth]{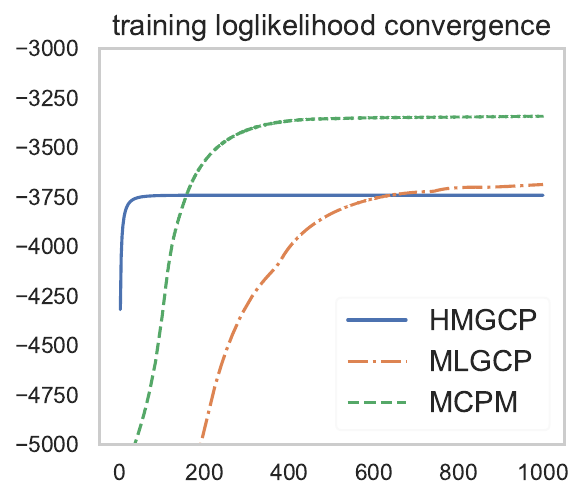}
\subcaption{Size of mask: $10\times10$}
\end{minipage}
\begin{minipage}{0.31\textwidth}
\centering
\includegraphics[width=\columnwidth]{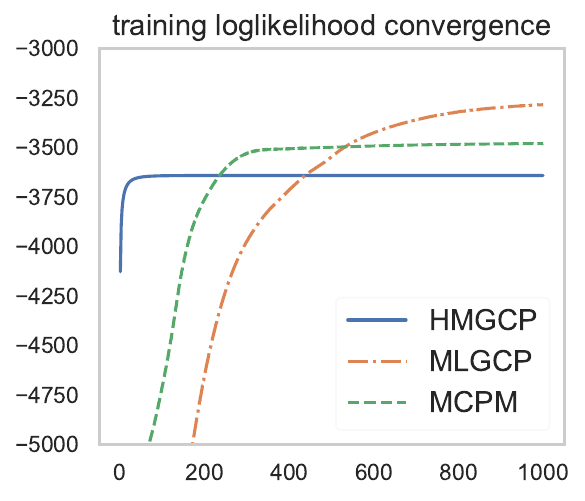}
\subcaption{Size of mask: $20\times20$}
\end{minipage}
\caption{The training log-likelihood convergence of HMGCP, MLGCP and MCPM. HMGCP only takes 40-50 steps to converge, while MCPM and MLGCP require more than 400 and 1000 steps to converge respectively. MLGCP and MCPM achieve the higher training log-likelihood due to overfitting.}
\label{app.fig5}
\end{center}
\end{figure}

We use $10\times5$ uniformly distributed inducing points horizontally and vertically on each task and $50\times25$ Gaussian quadrature nodes for the intractable integral. 
We randomly mask regions as explained above, and use the remaining data for training and the masked data for testing. 
\Cref{app.fig3} shows several examples of estimated latent functions from HMGCP with 3 different mask sizes (two examples for each size), while \cref{app.fig4} shows the corresponding estimated intensity functions from MLGCP and MCPM. 
The black boxes in~\cref{app.fig3} represent several possible configuration of masked regions on two Cox process tasks. It is easily observed in the data that in terms of income level and education degree, the west is significantly higher than the east; while for crime rate and non-market housing, it is the other way around. HMGCP successfully transfers knowledge existing in regression and classification tasks to help recover the intensity functions in masked regions for Cox process tasks (\cref{app.fig3}), while MLGCP and MCPM are prone to overfitting because they can only transfer homogeneous knowledge (\cref{app.fig4}). Therefore, HMGCP defeats the competing baselines MLGCP and MCPM in terms of TLL in all experiments (\cref{tab3}). More importantly, HMGCP has a faster convergence, which needs 40-50 steps to converge in terms of training log-likelihood, than the first-order gradient-based MLGCP and MCPM requiring more than 400 and 1000 steps respectively (\cref{app.fig5}). Besides, HMGCP significantly outperforms MLGCP and MCPM in terms of efficiency (\cref{tab3}, only on two Cox process tasks for a fair comparison). 

\begin{table}[t]
\caption{The performance of TLL and RT for HMGCP, MLGCP and MCPM on the real data over ten random configurations of masked regions with three different sizes of mask. The mean and standard deviation (in brackets) are provided. Time in seconds.}
\centering
\scalebox{0.8}{
\begin{sc}
\begin{tabular}{ccccc}
\toprule
Size of Mask & Model & TLL (crime) & TLL (non-market house) & RT (per step)\\
\midrule
\multirow{3}{*}{$5\times5$} & HMGCP & \textbf{-14.20}(12.14)  & \textbf{-14.22}(9.71) & \textbf{2.70}\\
& MLGCP & -22.71(22.67) & -23.67(21.67) & 7.82 \\
& MCPM & -24.40(23.32) & -20.08(13.93) & 12.02 \\
\midrule
\multirow{3}{*}{$10\times10$} & HMGCP & \textbf{-66.58}(28.91) & \textbf{-33.55}(22.90) & \textbf{2.67}\\
& MLGCP & -111.55(70.76) & -48.54(16.59) & 7.39 \\
& MCPM  & -115.18(64.25) & -47.76(14.12) & 11.81 \\
\midrule
\multirow{3}{*}{$20\times20$} & HMGCP & \textbf{-313.75}(133.26) & \textbf{-143.11}(82.89) & \textbf{2.49}\\
& MLGCP & -776.84(425.69) & -363.07(305.55) & 6.13 \\
& MCPM  & -558.02(205.73) & -223.67(101.72) & 11.85 \\
\bottomrule
\end{tabular}
\end{sc}}
\label{tab3}
\end{table}

\section{Conclusion}
The main objective of this study is to provide a heterogeneous multi-task learning framework for the analysis of multivariate inhomogeneous Poisson processes data with correlated regression and classification tasks. We adopt the MOGP prior to provide a shared representation to allow the transfer of knowledge between heterogeneous tasks. To circumvent the non-conjugate Bayesian inference, we employ the data augmentation technique to derive a closed-form mean-field approximation. Experimental results on synthetic and real data demonstrate that our model successfully shares the heterogeneous information to enhance the generalization capability and our inference approach has the predominant efficiency and convergence. 

We adopted the LMC based MOGP to incorporate the correlation between multiple heterogeneous tasks. An interesting research track in the future may be the extension to MOGP based on process convolution, which may bring more benefits on computation efficiency. Moreover, we only consider three kinds of heterogeneous tasks: regression, classification and Cox process in this work; other kinds of unsupervised tasks, such as clustering, can also be attempted to be introduced to the multi-task framework.

\backmatter




\begin{appendices}

\section{Proof of Augmented Likelihood for Classification}
\label{app.corollary1}
Substituting \cref{eq4} in the paper into the classification likelihood \cref{eq1b} in the paper, we can obtain
\begin{equation}
p(\mathbf{y}^{c}\mid\{g_{i}^c\}_{i=1}^{I_c})=\prod_{i=1}^{I_c}\prod_{n=1}^{N^c_i}\int_0^\infty e^{h(\omega^c_{i,n},y^c_{i,n}g^c_{i,n})}p_{\text{PG}}(\omega^c_{i,n}\mid1,0)d\omega^c_{i,n}, 
\label{app.eq1}
\end{equation}
where the integrand is the augmented likelihood:
\begin{equation}
p(\mathbf{y}^c,\bm{\omega}^c\mid\{g^c_i\}_{i=1}^{I_c})=\prod_{i=1}^{I_c}\prod_{n=1}^{N^c_i}e^{h(\omega^c_{i,n},y^c_{i,n}g^c_{i,n})}p_{\text{PG}}(\omega^c_{i,n}\mid1,0). 
\label{app.eq2}
\end{equation}

\section{Proof of Augmented Likelihood for Cox Process}
\label{app.corollary2}
Substituting \cref{eq4,eq6} in the paper into the product and exponential integral terms respectively in the Cox process likelihood \cref{eq1c} in the paper, we can obtain
\begin{equation}
\begin{aligned}
&p(\mathbf{x}^p\mid\{\bar{\lambda}_i,g^p_i\}_{i=1}^{I_p})=\prod_{i=1}^{I_p}\prod_{n=1}^{N_i^p}\int_0^\infty\Lambda_{i}(\mathbf{x}^p_{i,n},\omega^p_{i,n})e^{h(\omega^p_{i,n},g^p_{i,n})}d\omega^p_{i,n}\\
&\int_{\mathcal{X}}\int_0^\infty p_{\Lambda_i}(\Pi_i\mid\bar{\lambda}_i)\prod_{(\omega,\mathbf{x})\in\Pi_i}e^{h(\omega,-g^p_i(\mathbf{x}))}d\omega d\mathbf{x}, 
\end{aligned}
\label{app.eq3}
\end{equation}
where the integrand is the augmented likelihood:
\begin{equation}
\begin{aligned}
&p(\mathbf{x}^p,\bm{\omega}^p,\Pi\mid\bar{\bm{\lambda}}, \{g^p_i\}_{i=1}^{I_p})=\prod_{i=1}^{I_p}\prod_{n=1}^{N_i^p}\Lambda_{i}(\mathbf{x}^p_{i,n},\omega^p_{i,n})e^{h(\omega^p_{i,n},g^p_{i,n})}p_{\Lambda_i}(\Pi_i\mid\bar{\lambda}_i)\\
&\prod_{(\omega,\mathbf{x})\in\Pi_i}e^{h(\omega,-g^p_i(\mathbf{x}))}. 
\end{aligned}
\label{app.eq4}
\end{equation}

\section{Proof of Mean-Field Approximation}
\label{app.mean-field}
The augmented joint distribution can be written as:
\begin{equation}
\begin{aligned}
&p(\mathbf{y}^{r},\mathbf{y}^{c},\mathbf{x}^p,\bm{\omega}^c,\bm{\omega}^p,\Pi,g,\bar{\bm{\lambda}})\\
=&\underbrace{p(\mathbf{y}^{r}\mid\{g_{i}^r\}_{i=1}^{I_r})}_{\text{regression}}\underbrace{p(\mathbf{y}^c,\bm{\omega}^c\mid\{g^c_i\}_{i=1}^{I_c})}_{\text{augmented classification}}\underbrace{p(\mathbf{x}^p,\bm{\omega}^p,\Pi\mid\bar{\bm{\lambda}}, \{g^p_i\}_{i=1}^{I_p})}_{\text{augmented Cox process}}\underbrace{p(g)}_{\text{MOGP}}p(\bar{\bm{\lambda}})\\
=&\prod_{i=1}^{I_r}\prod_{n=1}^{N_{i}^r}\mathcal{N}(y^r_{i,n}\mid g^r_{i,n},\sigma_i^2)\prod_{i=1}^{I_c}\prod_{n=1}^{N^c_i}e^{h(\omega^c_{i,n},y^c_{i,n}g^c_{i,n})}p_{\text{PG}}(\omega^c_{i,n}\mid1,0)\\
&\prod_{i=1}^{I_p}\prod_{n=1}^{N_i^p}\Lambda_{i}(\mathbf{x}^p_{i,n},\omega^p_{i,n})e^{h(\omega^p_{i,n},g^p_{i,n})}p_{\Lambda_i}(\Pi_i\mid\bar{\lambda}_i)\prod_{(\omega,\mathbf{x})\in\Pi_i}e^{h(\omega,-g^p_i(\mathbf{x}))}p(g)p(\bar{\bm{\lambda}}).
\end{aligned}
\label{app.eq5}
\end{equation}
Here, we assume the variational posterior $q(\bm{\omega}^c,\bm{\omega}^p,\Pi,g,\bar{\bm{\lambda}})=q_1(\bm{\omega}^c,\bm{\omega}^p,\Pi)q_2(g,\bar{\bm{\lambda}})$. To minimize the KL divergence between variational posterior and true posterior, it can be proved that the optimal distribution of each factor is the expectation of the logarithm of the joint distribution taken over variables in the other factor~\citep{bishop2006pattern}:

\begin{equation}
\begin{aligned}
q_1^*(\bm{\omega}^c,\bm{\omega}^p,\Pi)&\propto e^{{\mathbb{E}_{q_2}[\log p(\mathbf{y}^{r},\mathbf{y}^{c},\mathbf{x}^p,\bm{\omega}^c,\bm{\omega}^p,\Pi,g,\bar{\bm{\lambda}})]}},\\
q_2^*(g,\bar{\bm{\lambda}})&\propto e^{{\mathbb{E}_{q_1}[\log p(\mathbf{y}^{r},\mathbf{y}^{c},\mathbf{x}^p,\bm{\omega}^c,\bm{\omega}^p,\Pi,g,\bar{\bm{\lambda}})]}}.
\end{aligned}
\label{app.eq6}
\end{equation}

Substituting \cref{app.eq5} into \cref{app.eq6}, we can obtain the optimal variational distributions. 
The process of deriving variational posteriors for $\bm{\omega}^c$, $\bm{\omega}^p$, $\Pi$, and $\bar{\bm{\lambda}}$ is similar to that in~\citet{donner2018efficient}. The primary distinction lies in the treatment of the latent function $g$. Further details are provided below. 

\paragraph{The optimal density for P\'{o}lya-Gamma latent variables}
The optimal variational posteriors of $\bm{\omega}^c$ and $\bm{\omega}^p$ are
\begin{equation}
\begin{aligned}
q_1(\bm{\omega}^c)=\prod_{i=1}^{I_c}\prod_{n=1}^{N_i^c}p_{\text{PG}}(\omega^c_{i,n}\mid1,\tilde{g}^c_{i,n}),\ \ \ \ q_1(\bm{\omega}^p)=\prod_{i=1}^{I_p}\prod_{n=1}^{N_i^p}p_{\text{PG}}(\omega^p_{i,n}\mid1,\tilde{g}^p_{i,n}),
\end{aligned}
\label{app.eq7}
\end{equation}
where $\tilde{g}^\cdot_{i,n}=\sqrt{\mathbb{E}[{g^\cdot_{i,n}}^2]}$ and we adopt the tilted P\'{o}lya-Gamma distribution $p_{\text{PG}}(\omega\mid b,c)\propto e^{-c^2\omega/2}p_{\text{PG}}(\omega\mid b,0)$~\citep{polson2013bayesian}. 

\paragraph{The optimal intensity for marked Poisson processes}
The derivation of optimal variational posterior of $\Pi=\{\Pi_i\}_{i=1}^{I_p}$ is challenging, so we provide some details below. After taking expectation, we can obtain
\begin{equation}
\begin{aligned}
q_1(\Pi_i)
=&\frac{p_{\tilde{\Lambda}_i}(\Pi_i\mid\bar{\lambda}^1_i)\prod_{(\omega,\mathbf{x})\in\Pi_i}e^{-\frac{\mathbb{E}[g^p_i(\mathbf{x})]}{2}-\frac{\mathbb{E}[{g^p_i(\mathbf{x})}^2]}{2}\omega-\log2}}{\iint p_{\tilde{\Lambda}_i}(\Pi_i\mid\bar{\lambda}^1_i)\prod_{(\omega,\mathbf{x})\in\Pi_i}e^{-\frac{\mathbb{E}[g^p_i(\mathbf{x})]}{2}-\frac{\mathbb{E}[{g^p_i(\mathbf{x})}^2]}{2}\omega-\log2}d\omega d\mathbf{x}}\\
=&p_{\tilde{\Lambda}_i}(\Pi_i\mid\bar{\lambda}^1_i)\prod_{(\omega,\mathbf{x})\in\Pi_i}e^{-\frac{\mathbb{E}[g^p_i(\mathbf{x})]}{2}-\frac{\mathbb{E}[{g^p_i(\mathbf{x})}^2]}{2}\omega-\log2}\\
&\exp{\left(\int_{\mathcal{X}}\int_0^\infty(1-e^{-\frac{\mathbb{E}[g^p_i(\mathbf{x})]}{2}-\frac{\mathbb{E}[{g^p_i(\mathbf{x})}^2]}{2}\omega-\log2})\bar{\lambda}^1_i p_{\text{PG}}(\omega\mid 1,0)d\omega d\mathbf{x}\right)}\\
=&\prod_{(\omega,\mathbf{x})\in\Pi_i}\bar{\lambda}^1_i p_{\text{PG}}(\omega\mid 1,0)e^{-\frac{\mathbb{E}[g^p_i(\mathbf{x})]}{2}-\frac{\mathbb{E}[{g^p_i(\mathbf{x})}^2]}{2}\omega-\log2}\\
&\exp{\left(-\int_{\mathcal{X}}\int_0^\infty\bar{\lambda}^1_i p_{\text{PG}}(\omega\mid 1,0)e^{-\frac{\mathbb{E}[g^p_i(\mathbf{x})]}{2}-\frac{\mathbb{E}[{g^p_i(\mathbf{x})}^2]}{2}\omega-\log2}d\omega d\mathbf{x}\right)},
\end{aligned}
\label{app.eq8}
\end{equation}
where $\bar{\lambda}^1_i=e^{\mathbb{E}[\log\bar{\lambda}_i]}$ and $\tilde{\Lambda}_i(\mathbf{x},\omega)=\bar{\lambda}^1_i p_{\text{PG}}(\omega\mid 1,0)$. The second line of \cref{app.eq8} used Campbell's theorem $\mathbb{E}_{\Pi_i}\left[\exp{\left( \sum_{(\mathbf{x},\omega)\in\Pi_i}h(\mathbf{x},\omega)\right)}\right]=\exp{\left[\iint\left(e^{ h(\mathbf{x},\omega)}-1\right)\tilde{\Lambda}_i(\mathbf{x},\omega)d\omega d\mathbf{x}\right]}$. It is easy to see the posterior intensity of $\Pi_i$ is
\begin{equation}
\begin{aligned}
\Lambda_i^1(\mathbf{x},\omega)&=\bar{\lambda}^1_i p_{\text{PG}}(\omega\mid 1,0)e^{-\frac{\mathbb{E}[g^p_i(\mathbf{x})]}{2}-\frac{\mathbb{E}[{g^p_i(\mathbf{x})}^2]}{2}\omega-\log2}\\
&=\bar{\lambda}_i^1 s(-\tilde{g}^p_i(\mathbf{x}))p_{\text{PG}}(\omega\mid 1,\tilde{g}^p_i(\mathbf{x}))e^{(\tilde{g}^p_i(\mathbf{x})-\bar{g}^p_i(\mathbf{x}))/2},
\end{aligned}
\label{app.eq9}
\end{equation}
where we adopt $e^{-c^2\omega/2}p_{\text{PG}}(\omega\mid b,0)=2s(-c)e^{c/2}p_{\text{PG}}(\omega\mid b,c)$~\citep{polson2013bayesian}, $\tilde{g}_i^p(\mathbf{x})=\sqrt{\mathbb{E}[{g_i^p(\mathbf{x})}^2]}$, $\bar{g}_i^p(\mathbf{x})=\mathbb{E}[g_i^p(\mathbf{x})]$. 

\paragraph{The optimal density for intensity upper-bounds}
The optimal variational posterior of $\bar{\bm{\lambda}}$ is
\begin{equation}
\begin{aligned}
q_2(\bar{\bm{\lambda}})=\prod_{i=1}^{I_p}p_{\text{Ga}}(\bar{\lambda}_i\mid N_i^p+R_i,\lvert\mathcal{X}\rvert),
\end{aligned}
\label{app.eq10}
\end{equation}
where $p_{\text{Ga}}$ is Gamma density, $R_i=\int_\mathcal{X}\int_0^\infty\Lambda_i^1(\mathbf{x},\omega)d\omega d\mathbf{x}$, $\lvert\mathcal{X}\rvert$ is the domain size. 

\paragraph{The optimal density for latent functions}
The derivation of optimal variational posterior of $g$ is challenging, so we provide some details below. After taking expectation, we can obtain
\begin{equation}
\begin{aligned}
&\log q_2(g)=\sum_{i=1}^{I_r}\sum_{n=1}^{N_i^r}\log\mathcal{N}(y_{i,n}^r\mid g_{i,n}^r,\sigma_i^2)+\sum_{i=1}^{I_c}\sum_{n=1}^{N_i^c}\left[\frac{y_{i.n}^c g_{i,n}^c}{2}-\frac{{g_{i,n}^c}^2}{2}\mathbb{E}[\omega_{i,n}^c]\right]+\\
&\sum_{i=1}^{I_p}\left[\sum_{n=1}^{N_i^p}\frac{g_{i,n}^p}{2}-\frac{{g_{i,n}^p}^2}{2}\mathbb{E}[\omega_{i,n}^p]-\mathbb{E}_{\Pi_i}\sum_{(\omega,\mathbf{x})\in\Pi_i}\frac{g_i^p(\mathbf{x})}{2}+\frac{{g_i^p(\mathbf{x})}^2}{2}\omega\right]+\log p(g)+C\\
&=\sum_{i=1}^{I_r}\sum_{n=1}^{N_i^r}\log\mathcal{N}(g_{i,n}^r\mid y_{i,n}^r,\sigma_i^2)+\sum_{i=1}^{I_c}\sum_{n=1}^{N_i^c}\log\mathcal{N}(g_{i,n}^c\mid \frac{y_{i,n}^c}{2\mathbb{E}[\omega_{i,n}^c]},\frac{1}{\mathbb{E}[\omega_{i,n}^c]})\\
&+\sum_{i=1}^{I_p}\left[\int_{\mathcal{X}}\sum_{n=1}^{N_i^p}\left(\frac{g_{i}^p(\mathbf{x})}{2}-\frac{{g_{i}^p(\mathbf{x})}^2}{2}\mathbb{E}[\omega_{i,n}^p]\right)\delta(\mathbf{x}-\mathbf{x}^p_{i,n})d\mathbf{x}\right.\\
&\left.-\int_{\mathcal{X}}\int_0^\infty\left(\frac{g_i^p(\mathbf{x})}{2}+\frac{{g_i^p(\mathbf{x})}^2}{2}\omega\right)\Lambda_i^1(\mathbf{x},\omega)d\omega d\mathbf{x}\right]+\log p(g)+C\\
&=\sum_{i=1}^{I_r}\sum_{n=1}^{N_i^r}\log\mathcal{N}(g_{i,n}^r\mid y_{i,n}^r,\sigma_i^2)+\sum_{i=1}^{I_c}\sum_{n=1}^{N_i^c}\log\mathcal{N}(g_{i,n}^c\mid \frac{y_{i,n}^c}{2\mathbb{E}[\omega_{i,n}^c]},\frac{1}{\mathbb{E}[\omega_{i,n}^c]})\\
&+\sum_{i=1}^{I_p}\left[\int_{\mathcal{X}}\left(\frac{1}{2}\sum_{n=1}^{N^p_i}\delta(\mathbf{x}-\mathbf{x}^p_{i,n})-\frac{1}{2}\int_0^\infty\Lambda_i^1(\mathbf{x},\omega)d\omega\right)g_i^p(\mathbf{x})d\mathbf{x}+\log p(g)\right.\\
&\left.-\frac{1}{2}\int_{\mathcal{X}}\left(\sum_{n=1}^{N_i^p}\mathbb{E}[\omega_{i,n}^p]\delta(\mathbf{x}-\mathbf{x}^p_{i,n})+\int_0^\infty\omega\Lambda_i^1(\mathbf{x},\omega)d\omega\right){g_i^p(\mathbf{x})}^2 d\mathbf{x}\right]+C\\
&=\sum_{i=1}^{I_r}\sum_{n=1}^{N_i^r}\log\mathcal{N}(g_{i,n}^r\mid y_{i,n}^r,\sigma_i^2)+\sum_{i=1}^{I_c}\sum_{n=1}^{N_i^c}\log\mathcal{N}(g_{i,n}^c\mid \frac{y_{i,n}^c}{2\mathbb{E}[\omega_{i,n}^c]},\frac{1}{\mathbb{E}[\omega_{i,n}^c]})\\
&+\sum_{i=1}^{I_p}\left[\int_{\mathcal{X}} B_i(\mathbf{x})g_i^p(\mathbf{x})d\mathbf{x}-\frac{1}{2}\int_{\mathcal{X}} A_i(\mathbf{x}){g_i^p(\mathbf{x})}^2 d\mathbf{x}\right]+\log p(g)+C,
\end{aligned}
\label{app.eq11}
\end{equation}
where $A_i(\mathbf{x})=\sum_{n=1}^{N^p_i}\mathbb{E}[\omega_{i,n}^p]\delta(\mathbf{x}-\mathbf{x}^p_{i,n})+\int_0^\infty\omega\Lambda_i^1(\mathbf{x},\omega)d\omega$ and $B_i(\mathbf{x})=\frac{1}{2}\sum_{n=1}^{N^p_i}\delta(\mathbf{x}-\mathbf{x}^p_{i,n})-\frac{1}{2}\int_0^\infty\Lambda_i^1(\mathbf{x},\omega)d\omega$. 

The computation of \cref{app.eq11} suffers from a cubic complexity w.r.t. the number of data points in regression, classification and point process tasks. We use the inducing inputs formalism to make the inference scalable. We denote $M$ inducing inputs $[\mathbf{x}_1\,\ldots,\mathbf{x}_M]^\top$ on the domain $\mathcal{X}$ for each task. The function values of basis function $f_q$ at these inducing inputs are defined as $\mathbf{f}_{q,\mathbf{x}_m}$. Then we can obtain the function values of task-specific latent function $g_i$ at these inducing inputs $\mathbf{g}_{\mathbf{x}_m}^i=\sum_{q=1}^Qw_{i,q}\mathbf{f}_{q,\mathbf{x}_m}$. If we define $\mathbf{g}_{\mathbf{x}_m}=[\mathbf{g}^{\top}_{1,\mathbf{x}_m},\ldots,\mathbf{g}^{\top}_{I,\mathbf{x}_m}]^\top$, $\mathbf{g}_{\mathbf{x}_m}\sim\mathcal{N}(\mathbf{0},\mathbf{K}_{\mathbf{x}_m \mathbf{x}_m})$ where $\mathbf{K}_{\mathbf{x}_m\mathbf{x}_m}$ is the MOGP covariance on $\mathbf{x}_m$ for all tasks and $\mathbf{g}_{\mathbf{x}_m}^i\sim\mathcal{N}(\mathbf{0},\mathbf{K}_{\mathbf{x}_m \mathbf{x}_m}^i)$ where $\mathbf{K}_{\mathbf{x}_m\mathbf{x}_m}^i$ is $i$-th diagonal block of $\mathbf{K}_{\mathbf{x}_m\mathbf{x}_m}$. Given $\mathbf{g}_{\mathbf{x}_m}^i$, we assume the function $g_i(\mathbf{x})$ is the posterior mean function $g_i(\mathbf{x})=\mathbf{k}_{\mathbf{x}_m \mathbf{x}}^{i\top}\mathbf{K}_{\mathbf{x}_m \mathbf{x}_m}^{i^{-1}}\mathbf{g}_{\mathbf{x}_m}^i$ where $\mathbf{k}_{\mathbf{x}_m \mathbf{x}}^{i}$ is the kernel w.r.t. inducing points and predictive points for $i$-th task. Therefore, $\{g_{i,n}^r\}_{i=1}^{I_r}$, $\{g_{i,n}^c\}_{i=1}^{I_c}$ and $\{g_i^p(\mathbf{x})\}_{i=1}^{I_p}$ can be written as 
\begin{equation}
\begin{aligned}
\mathbf{g}_{i}^r=\mathbf{K}_{\mathbf{x}_m \mathbf{x}_n}^{r,i\top}\mathbf{K}_{\mathbf{x}_m \mathbf{x}_m}^{r,i^{-1}}\mathbf{g}_{\mathbf{x}_m}^{r,i},\mathbf{g}_{i}^c=\mathbf{K}_{\mathbf{x}_m \mathbf{x}_n}^{c,i\top}\mathbf{K}_{\mathbf{x}_m \mathbf{x}_m}^{c,i^{-1}}\mathbf{g}_{\mathbf{x}_m}^{c,i},
g_i^p(\mathbf{x})=\mathbf{k}_{\mathbf{x}_m \mathbf{x}}^{p,i\top}\mathbf{K}_{\mathbf{x}_m \mathbf{x}_m}^{p,i^{-1}}\mathbf{g}_{\mathbf{x}_m}^{p,i}, 
\end{aligned}
\label{app.eq12}
\end{equation}
where $\mathbf{g}_{i}^r=[g_{i,1}^r,\ldots,g_{i,N_i^r}^r]^\top$, $\mathbf{g}_{i}^c=[g_{i,1}^c,\ldots,g_{i,N_i^c}^c]^\top$, $g_i^p(\mathbf{x})$ is the function value of $g_i^p$ on $\mathbf{x}$. 

Substituting \cref{app.eq12} into \cref{app.eq11}, we obtain the inducing points version of \cref{app.eq11}:
\begin{equation}
\begin{aligned}
&q_2(\mathbf{g}_{\mathbf{x}_m})\propto\prod_{i=1}^{I_r}\mathcal{N}(\mathbf{K}_{\mathbf{x}_m \mathbf{x}_n}^{r,i\top}\mathbf{K}_{\mathbf{x}_m \mathbf{x}_m}^{r,i^{-1}}\mathbf{g}_{\mathbf{x}_m}^{r,i}\mid\mathbf{y}_i^r,\text{diag}(\sigma_i^2))\\
&\cdot\prod_{i=1}^{I_c}\mathcal{N}(\mathbf{K}_{\mathbf{x}_m \mathbf{x}_n}^{c,i\top}\mathbf{K}_{\mathbf{x}_m \mathbf{x}_m}^{c,i^{-1}}\mathbf{g}_{\mathbf{x}_m}^{c,i}\mid \frac{\mathbf{y}_{i}^c}{2\mathbb{E}[\bm{\omega}_{i}^c]},\text{diag}(\frac{1}{\mathbb{E}[\bm{\omega}_{i}^c]}))\\
&\cdot\prod_{i=1}^{I_p}\exp\left(\int_{\mathcal{X}} B_i(\mathbf{x})\mathbf{k}_{\mathbf{x}_m \mathbf{x}}^{p,i\top} d\mathbf{x}\mathbf{K}_{\mathbf{x}_m \mathbf{x}_m}^{p,i^{-1}}\mathbf{g}_{\mathbf{x}_m}^{p,i}\right.\\
&\left.-\frac{1}{2}\mathbf{g}_{\mathbf{x}_m}^{p,i\top}\mathbf{K}_{\mathbf{x}_m \mathbf{x}_m}^{p,i^{-1}}\int_{\mathcal{X}} A_i(\mathbf{x})\mathbf{k}_{\mathbf{x}_m \mathbf{x}}^{p,i}\mathbf{k}_{\mathbf{x}_m \mathbf{x}}^{p,i\top} d\mathbf{x}\mathbf{K}_{\mathbf{x}_m \mathbf{x}_m}^{p,i^{-1}}\mathbf{g}_{\mathbf{x}_m}^{p,i}\right)\\
&\cdot\mathcal{N}(\mathbf{g}_{\mathbf{x}_m}\mid\mathbf{0},\mathbf{K}_{\mathbf{x}_m \mathbf{x}_m}). 
\end{aligned}
\label{app.eq13}
\end{equation}

It is easy to see the third line of \cref{app.eq13} is a multivariate Gaussian distribution of $\mathbf{g}_{\mathbf{x}_m}^{p,i}$. The likelihoods of $\mathbf{g}_{\mathbf{x}_m}^{r,i}$ for regression, $\mathbf{g}_{\mathbf{x}_m}^{c,i}$ for classification and $\mathbf{g}_{\mathbf{x}_m}^{p,i}$ for point process tasks are all Gaussian distributions, so they are conjugate to the MOGP prior and we can obtain the closed-form variational posterior for $\mathbf{g}_{\mathbf{x}_m}$:
\begin{equation}
\begin{gathered}
q_2(\mathbf{g}_{\mathbf{x}_m})=\mathcal{N}(\mathbf{g}_{\mathbf{x}_m}\mid\mathbf{m}_{\mathbf{x}_m},\mathbf{\Sigma}_{\mathbf{x}_m}),
\end{gathered}
\label{app.eq14}
\end{equation}
where $\mathbf{g}_{\mathbf{x}_m}=[\mathbf{g}_{\mathbf{x}_m}^{r\top},\mathbf{g}_{\mathbf{x}_m}^{c\top},\mathbf{g}_{\mathbf{x}_m}^{p\top}]^\top$, $\mathbf{g}_{\mathbf{x}_m}^{\cdot}=[\mathbf{g}_{1,\mathbf{x}_m}^{\cdot\top},\ldots,\mathbf{g}_{I_\cdot,\mathbf{x}_m}^{\cdot\top}]^\top$ and 
\begin{equation*}
\mathbf{\Sigma}_{\mathbf{x}_m}=\left[\text{diag}\left(\mathbf{H}_{\mathbf{x}_m}^r,\mathbf{H}_{\mathbf{x}_m}^c,\mathbf{H}_{\mathbf{x}_m}^p\right)+\mathbf{K}_{\mathbf{x}_m \mathbf{x}_m}^{-1}\right]^{-1},\mathbf{m}_{\mathbf{x}_m}=\mathbf{\Sigma}_{\mathbf{x}_m}[\mathbf{v}_{\mathbf{x}_m}^{r\top},\mathbf{v}_{\mathbf{x}_m}^{c\top},\mathbf{v}_{\mathbf{x}_m}^{p\top}]^\top,
\end{equation*}
where $\mathbf{H}_{\mathbf{x}_m}^\cdot=\text{diag}(\mathbf{H}_{1,\mathbf{x}_m}^\cdot,\ldots,\mathbf{H}_{I_\cdot,\mathbf{x}_m}^\cdot)$, $\mathbf{v}_{\mathbf{x}_m}^\cdot=[\mathbf{v}_{1,\mathbf{x}_m}^{\cdot\top},\ldots,\mathbf{v}_{I_\cdot,\mathbf{x}_m}^{\cdot\top}]^\top$ and 
\begin{equation*}
\begin{gathered}
\mathbf{H}_{i,\mathbf{x}_m}^r=\mathbf{K}_{\mathbf{x}_m \mathbf{x}_m}^{r,i^{-1}}\mathbf{K}_{\mathbf{x}_m \mathbf{x}_n}^{r,i}\mathbf{D}^r_i\mathbf{K}_{\mathbf{x}_m \mathbf{x}_n}^{r,i\top}\mathbf{K}_{\mathbf{x}_m \mathbf{x}_m}^{r,i^{-1}},
\mathbf{v}_{i,\mathbf{x}_m}^{r}=\mathbf{K}_{\mathbf{x}_m \mathbf{x}_m}^{r,i^{-1}}\mathbf{K}_{\mathbf{x}_m \mathbf{x}_n}^{r,i}\frac{\mathbf{y}^r_i}{\sigma_i^2},\\
\mathbf{H}_{i,\mathbf{x}_m}^c=\mathbf{K}_{\mathbf{x}_m \mathbf{x}_m}^{c,i^{-1}}\mathbf{K}_{\mathbf{x}_m \mathbf{x}_n}^{c,i}\mathbf{D}^c_i\mathbf{K}_{\mathbf{x}_m \mathbf{x}_n}^{c,i\top}\mathbf{K}_{\mathbf{x}_m \mathbf{x}_m}^{c,i^{-1}},
\mathbf{v}_{i,\mathbf{x}_m}^{c}=\mathbf{K}_{\mathbf{x}_m \mathbf{x}_m}^{c,i^{-1}}\mathbf{K}_{\mathbf{x}_m \mathbf{x}_n}^{c,i}\frac{\mathbf{y}^c_i}{2},\\
\mathbf{H}_{i,\mathbf{x}_m}^p=\mathbf{K}_{\mathbf{x}_m \mathbf{x}_m}^{p,i^{-1}}\int_{\mathcal{X}}A_i(\mathbf{x})\mathbf{k}_{\mathbf{x}_m \mathbf{x}}^{p,i}\mathbf{k}_{\mathbf{x}_m \mathbf{x}}^{p,i\top}d\mathbf{x}\mathbf{K}_{\mathbf{x}_m \mathbf{x}_m}^{p,i^{-1}},\\
\mathbf{v}_{i,\mathbf{x}_m}^{p}=\mathbf{K}_{\mathbf{x}_m \mathbf{x}_m}^{p,i^{-1}}\int_{\mathcal{X}}B_i(\mathbf{x})\mathbf{k}_{\mathbf{x}_m \mathbf{x}}^{p,i}d\mathbf{x},
\end{gathered}
\end{equation*}
where $\mathbf{D}^r_i=\text{diag}(1/\sigma_i^2)$ and $\mathbf{D}^c_i=\text{diag}(\mathbb{E}[\bm{\omega}^{c}_i])$. 

\section{Multi-class Classification}
\label{app.multiclass}
In the paper, we mainly focus on the binary classification problem because each binary classification task corresponds to a single latent function. This setting is consistent with the regression and point process tasks in which each task only specifies a single latent function. 

For $Z$-class classification problem, each task corresponds to $Z$ latent functions. The usual likelihood for multi-class classification is the softmax function:
\begin{equation}
\begin{aligned}
p(y_{i,n}^c=k\mid\mathbf{f}_{i,n}^c)=\frac{e^{(f_{i,n}^{c,k})}}{\sum_{z=1}^Ze^{(f_{i,n}^{c,z})}},
\end{aligned}
\label{app.eq18}
\end{equation}
where $f_{i,n}^{c,k}=f_{i}^{c,k}(\mathbf{x}_n)$, $\mathbf{f}_{i,n}^c=[f_{i,n}^{c,1},\ldots,f_{i,n}^{c,Z}]^{\top}$, $k\in\{1,\ldots,Z\}$. However, the P\'{o}lya-Gamma augmentation technique for binary classification can not be directly employed in the softmax function. \citet{galy2020multi} and \citet{jake2021bayesian} proposed the \emph{logistic-softmax function} and the \emph{one-vs-each softmax approximation} respectively that enable us to employ P\'{o}lya-Gamma augmentation to obtain a conditionally conjugate model for multi-class classification tasks. Both methods mentioned above can be incorporated into our framework in the multi-class classification scenario. We refer the readers to \citet{galy2020multi,jake2021bayesian} for more details.

\section{Comparison with HetMOGP}

One anonymous reviewer point out that an important baseline to compare against is \citet{moreno2018heterogeneous} that can also handle regression, classification and counting data, even if the discretized Poisson distribution likelihood is used instead of the continuous point process likelihood considered in this work. \Citet{moreno2018heterogeneous} used the generic variational inference method mentioned in the introduction for parameter posterior, so this comparison can demonstrate the advantage of using data augmentation for conjugate operations. 

We compare the performance of TLL and RT for HMGCP and heterogeneous multi-output Gaussian process (HetMOGP)~\citep{moreno2018heterogeneous} on the synthetic data from \cref{sec5.1,sec5.2}. 
Since HetMOGP can only handle discrete count data, we discretize the original observation window $[0,100]$ into $100$ bins and then calculate the number of points in each bin separately. 
We use the default hyperparameter settings in the demo code provided by~\citet{moreno2018heterogeneous}. 
The results are shown in \cref{app.tab1,app.tab2}. 
From \cref{app.tab1,app.tab2}, we can see that HMGCP has the better TLL than HetMOGP that is trained on the discrete count data.  
For a fair comparison of efficiency, we run both HMGCP and HetMOGP on all tasks, and our inference is much faster than HetMOGP. 
This is because, for HetMOGP, it uses the generic variational inference, so the numerical optimization has to be performed during the variational iterations; while for our model HMGCP, the variational iterations have completely analytical expressions due to data augmentation, so it leads to the more efficient computation. 
It is worth noting that the running times presented in \cref{app.tab1,app.tab2} encompass all tasks (regression, classification, and Cox processes), resulting in longer duration compared to those reported in \cref{sec5.1,sec5.2}, which are solely based on the Cox process tasks.

\begin{table}[t]
\caption{The performance of TLL and RT for HMGCP and HetMOGP on three synthetic datasets in Section 5.1. Time in seconds.}
\label{app.tab1}
\begin{center}
\scalebox{0.7}{
\begin{sc}
\begin{tabular}{lccccc}
\toprule
& Model & TLL(reg) & TLL(cla) & TLL(Cox) & RT (400 iterations)\\
\midrule
\multirow{2}{*}{1} & HMGCP & \textbf{-33.17} & \textbf{-63.57} & \textbf{-89.05} & \textbf{1.6}\\
& HetMOGP & -97.80 & -66.22 & -181.91 & 708.74 \\
\midrule
\multirow{2}{*}{2} & HMGCP & \textbf{-28.54} & \textbf{-55.23} & \textbf{-63.54} & \textbf{1.79}\\
& HetMOGP & -98.8 & -58.1 & -196.71 & 812.88 \\
\midrule
\multirow{2}{*}{3} & HMGCP & \textbf{-42.43} & \textbf{-56.14} & \textbf{-72.75} & \textbf{1.54}\\
& HetMOGP & -138.21 & -65.01 & -172.77 & 647.70 \\
\bottomrule
\end{tabular}
\end{sc}}
\end{center}
\end{table}

\begin{table}[t]
\caption{The performance of TLL and RT for HMGCP and HetMOGP on synthetic datasets in Section 5.2 over ten random configurations of missing gaps with three different missing-gap widths ($0$ means complete data). The mean and standard deviation (in brackets) are provided. TLL(Cox) is the sum of TLLs of two Cox processes. Time in seconds.}
\label{app.tab2}
\centering
\scalebox{0.7}{
\begin{sc}
\begin{tabular}{cccccc}
\toprule
Gap Width & Model & TLL(reg) & TLL(cla) & TLL(Cox) & RT (2000 iterations)\\
\midrule
\multirow{2}{*}{0} & HMGCP & \textbf{-50.61} & \textbf{-56.67} & \textbf{-120.55} & \textbf{12.9}\\

& HetMOGP &  -101.19 & -64.18 & -380.63 & 4029.75 \\
\midrule
\multirow{2}{*}{5} & HMGCP & \textbf{-50.76}(0.92) & \textbf{-56.74}(0.51) & \textbf{-122.94}(2.27) & \textbf{12.5}\\
& HetMOGP & -105.19 (2.92) & -73.99 (12.01) & -391.38 (29.00)  &  3826.56\\
\midrule
\multirow{2}{*}{10} & HMGCP & \textbf{-52.14}(1.94) & \textbf{-56.82}(0.69) & \textbf{-128.49}(5.74) & \textbf{11.7}\\
& HetMOGP & -104.45 (10.49) &  -66.12 & -414.58 (21.78) &  3424.73\\
\bottomrule
\end{tabular}
\end{sc}}
\end{table}

\end{appendices}

\bibliography{sn-bibliography}


\end{document}